%% file: main.tex
\documentclass[10pt,twocolumn,letterpaper]{article}

\usepackage[pagenumbers]{cvpr} % To force page numbers, e.g. for an arXiv version

\usepackage{graphicx}
\usepackage{amsmath}
\usepackage{amssymb}
\usepackage{booktabs}
\usepackage{xcolor}
\usepackage{multirow}
\usepackage{array}
\usepackage{pifont}
\usepackage{tabularray}
\usepackage{algorithm2e}

\newcommand{\xmark}{\text{\ding{55}}}
\usepackage[pagebackref,breaklinks,colorlinks]{hyperref}

\usepackage[capitalize]{cleveref}
\crefname{section}{Sec.}{Secs.}
\Crefname{section}{Section}{Sections}
\Crefname{table}{Table}{Tables}
\crefname{table}{Tab.}{Tabs.}

\makeatletter\renewcommand\paragraph{\@startsection{paragraph}{4}{\z@}
  {.5em \@plus1ex \@minus.2ex}{-.5em}{\normalfont\normalsize\bfseries}}\makeatother

\newcommand{\zc}[1]{{\color{blue}[ZC: #1]}}

\def\cvprPaperID{624} % *** Enter the CVPR Paper ID here
\def\confName{CVPR}
\def\confYear{2023}

\begin{document}

% \title{PolyFormer: A Unified Model for Referring Image Segmentation}
\title{PolyFormer: Referring Image Segmentation as Sequential Polygon Generation}
\author{
    Jiang Liu$^{1,\dag,}$\thanks{Work done during internship at AWS AI. $\dag$ Equal contribution.} 
    \hspace{0.3cm} Hui Ding$^{2,\dag}$
    \hspace{0.3cm} Zhaowei Cai$^{2}$
    \hspace{0.3cm} Yuting Zhang$^{2}$ \\
    \hspace{0.3cm} Ravi Kumar Satzoda$^{2}$
    \hspace{0.3cm} Vijay Mahadevan$^{2}$
    \hspace{0.3cm} R. Manmatha$^{2}$ \\ [.5ex]
    Johns Hopkins University$^{1}$ \hspace{0.3cm} AWS AI Labs$^{2}$ \\ [.5ex]
    % \url{https://polyformer.github.io/}
    % {\tt\small jiangliu@jhu.edu, \{huidin,zhaoweic,yutingzh,ravisatz,vmahad,manmatha\}@amazon.com}
    {\tt\small \url{https://polyformer.github.io/}}
}
\maketitle

%%%%%%%%% ABSTRACT
\vspace{-4mm}
\begin{abstract}

In this work, instead of directly predicting the pixel-level segmentation masks, the problem of referring image segmentation is formulated as sequential polygon generation, and the predicted polygons can be later converted into segmentation masks. This is enabled by a new sequence-to-sequence framework, Polygon Transformer (PolyFormer), which takes a sequence of image patches and text query tokens as input, and outputs a sequence of polygon vertices autoregressively. For more accurate geometric localization, we propose a regression-based decoder, which predicts the precise floating-point coordinates directly, without any coordinate quantization error. In the experiments,  PolyFormer outperforms the prior art by a clear margin, e.g., 5.40\% and 4.52\% absolute improvements on the challenging RefCOCO+ and RefCOCOg datasets. It also shows strong generalization ability when evaluated on the referring video segmentation task without fine-tuning, e.g., achieving competitive 61.5\% $\mathcal{J}\&\mathcal{F}$ on the Ref-DAVIS17 dataset. 
% Our project page is at \url{https://polyformer.github.io/}.

\end{abstract}

\setlength{\abovedisplayskip}{5pt}
\setlength{\belowdisplayskip}{5pt}
\setlength{\abovecaptionskip}{5pt}
\setlength{\belowcaptionskip}{-2pt}

%%%%%%%%% BODY TEXT
\vspace{-2mm}
\section{Introduction}
\label{sec:intro}
Referring image segmentation (RIS)  \cite{liu2017recurrent,li2018referring,margffoy2018dynamic,shi2018key,chen2019see,ye2019cross,hu2020bi,hui2020linguistic,huang2020referring,feng2021encoder,ding2021vlt,yang2022lavt,wang2022cris,kim2022restr, hu2016segmentation} combines vision-language understanding \cite{lu2019vilbert,tan2019lxmert,lu202012,zhang2021vinvl,li2021align} and instance segmentation \cite{he2017mask, liu2018path, chen2019hybrid, dai2016instance,bolya2019yolact}, and aims to localize the segmentation mask of an object given a natural language query. It generalizes traditional object segmentation from a fixed number of predefined categories to any concept described by free-form language, which requires a deeper understanding of the image and language semantics. The conventional pipeline~\cite{liu2017recurrent,li2018referring,margffoy2018dynamic,shi2018key,chen2019see,ye2019cross,hu2020bi,hui2020linguistic,huang2020referring,feng2021encoder,ding2021vlt, hu2016segmentation} first extracts features from the image and text inputs, and then fuses the multi-modal features together to predict the mask.

%Most instance segmentation models \cite{he2017mask, liu2018path, chen2019hybrid, dai2016instance} rely on a dense binary classification network to determine whether each pixel belongs to a referred object. This spatial pixel-wise prediction output by a convolutional network neglects the structure among the output predictions. For example, each pixel is predicted independently of other pixels. 

A segmentation mask encodes the spatial layout of an object, and most instance segmentation models \cite{he2017mask, liu2018path, chen2019hybrid, dai2016instance} rely on a dense binary classification network to determine whether each pixel belongs to the object. This pixel-to-pixel prediction is preferred by a convolutional operation, but it neglects the structure among the output predictions. For example, each pixel is predicted independently of other pixels. 
In contrast, a segmentation mask can also be represented by a sparse set of structured polygon vertices delineating the contour of the object \cite{mscoco,xie2020polarmask, castrejon2017annotating, polytransform, acuna2018efficient, boundaryformer}. This structured sparse representation is cheaper than a dense mask representation and is the preferred annotation format for most instance segmentation datasets \cite{mscoco, Cordts_2016_CVPR, russell2008labelme}. Thus, it is also tempting to predict structured polygons directly. 
\iffalse
It is also tempting to predict structured polygons directly. 
\fi
However, how to effectively predict this type of structured outputs is challenging, especially for convolutional neural networks (CNNs), and previous efforts have not shown much success yet \cite{xie2020polarmask, polytransform, boundaryformer}.

\begin{figure*}[t]
  \centering
\includegraphics[width=0.8\textwidth]{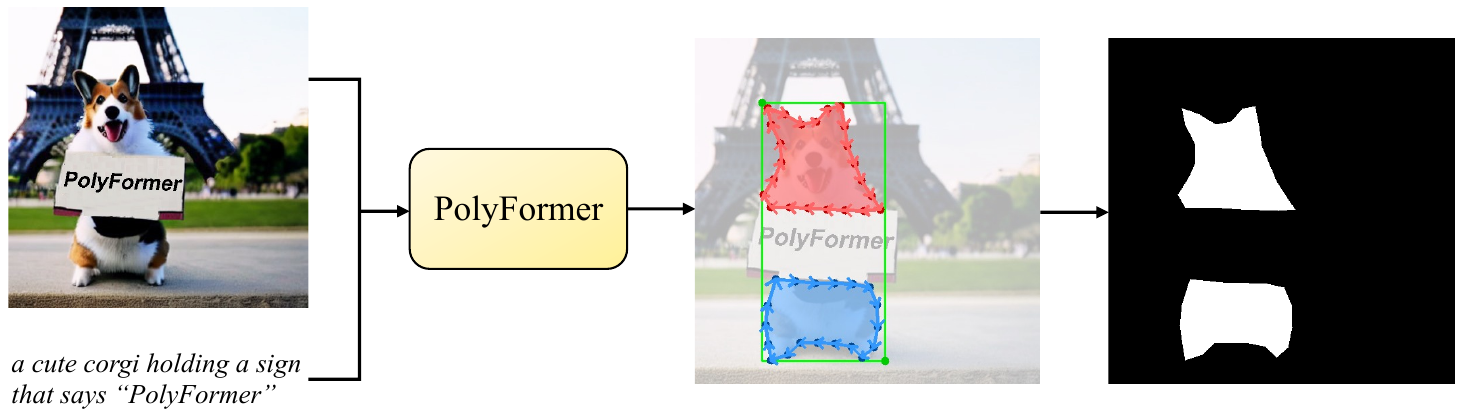}
  \caption{
%   \iffalse
%   We present a model called PolyFormer to generate a sequence of vertices outlining the referred object (red and blue dots). Based on the predicted polygons, it is straightforward to obtain the segmentation mask. Since a bounding box may also be represented as a sequence of two corner points (green dots),
%   PolyFormer unifies the referring image segmentation and
%   referring expression comprehension tasks in a single framework.
%   \fi
The illustration of PolyFormer pipeline for referring image segmentation (polygon vertex sequence) and referring expression comprehension (bounding box corner points). The polygons are converted to segmentation masks in the end.
  %We present a model called PolyFormer to sequentially generate the continuous coordinates of bounding box corner points (green dots) and vertices of polygons (red and blue dots) outlining the object in a unified framework. 
%   \zc{the arrows are not clear?}
%   We present a model called PolyFormer to generate a sequence of vertices outlining the referred object, here ``cute corgi". Due to the occlusion caused by the sign, two polygons (red and blue dots) are needed to describe the object. Based on the predicted polygons, it is straightforward to obtain the segmentation mask. Since a bounding box may also be represented as a sequence of two corner points (green dots),
%   PolyFormer unifies the referring image segmentation and
%   referring expression comprehension tasks in a single framework.
  }
  \label{fig:intro}
\end{figure*}

We address this challenge by resorting to a sequence-to-sequence (seq2seq) framework \cite{seq2seq, seq2seq2, radford2019language,raffel2020exploring,brown2020language}, and propose \textbf{Poly}gon trans\textbf{Former} (PolyFormer) for referring image segmentation. As illustrated in \cref{fig:intro}, it takes a sequence of image patches and text query tokens as input, and autoregressively outputs a sequence of polygon vertices. Since each vertex prediction is conditioned on all preceding predicted vertices, the output predictions are no longer independent of each other. 
The seq2seq framework is flexible on its input and output format, as long as both of them can be formulated as sequences of variable length. Thus, it is natural to concatenate the visual and language features together as a long sequence, avoiding complicated multi-modal feature fusion as in prior work \cite{shi2018key, chen2019see, ye2019cross, hu2020bi, yang2022lavt}. In the meantime, the output can also be a long sequence of multiple polygons separated by separator tokens, covering the scenario where the segmentation masks are not connected, \eg, by occlusion, as shown in Fig. \ref{fig:intro}. Furthermore, since a bounding box can be represented as a sequence of two corner points (\ie, top left and bottom right), they can also be output by PolyFormer along with the polygon vertices. Thus, referring image segmentation (polygon) and referring expression comprehension (bounding box) can be unified in our simple PolyFormer framework.

Localization is important for polygon and bounding box generation, as a single coordinate prediction mistake may result in substantial errors in the mask or bounding box prediction. However, in recent seq2seq models for the visual domain \cite{chen2021pix2seq,chen2022unified,wang2022unifying,lu2022unifiedio}, in order to accommodate all tasks in a unified seq2seq framework, the coordinates are quantized into discrete bins and the prediction is formulated as a classification task. This is not ideal since geometric coordinates lie in a continuous space instead of a discrete one, and classification is thus usually suboptimal for localization task \cite{DBLP:conf/cvpr/GidarisK16,li2018brute,zhou2019objects}. Instead, we formulate localization as a regression task, due to its success in object detection \cite{girshick2015fast,ren2015faster,he2017mask,cai2018cascade}, where floating-point coordinates are directly predicted without any quantization error. Motivated by \cite{he2017mask}, the feature embedding for any floating-point coordinate in PolyFormer is obtained by bilinear interpolation \cite{DBLP:conf/nips/JaderbergSZK15} of its neighboring indexed embeddings. This is in contrast with the common practice \cite{chen2021pix2seq,wang2022unifying,lu2022unifiedio} in which the coordinate feature is indexed from a dictionary with a fixed number of discrete coordinate bins. 
These changes enable our PolyFormer to make accurate polygon and bounding box predictions.

We evaluate PolyFormer on three major referring image segmentation benchmarks. 
It achieves 76.94\%, 72.15\%, and 71.15\% mIoU on the validation sets of RefCOCO~\cite{yu2016modeling}, RefCOCO+~\cite{yu2016modeling} and RefCOCOg~\cite{mao2016generation}, outperforming the state of the art by absolute margins of \textbf{2.48}\%, \textbf{5.40}\%, and \textbf{4.52}\%, respectively.
PolyFormer also shows strong generalization ability when directly applied to the referring video segmentation task without finetuning. It achieves 61.5\% $\mathcal{J}\&\mathcal{F}$ on the Ref-DAVIS17 dataset~\cite{khoreva2018video}, comparable with~\cite{wu2022referformer} which is specifically designed for that task.

Our main contributions are summarized as follows:
\vspace{-2mm}
\begin{itemize}
    %\item We introduce a novel seq2seq framework for referring image segmentation, called PolyFormer, which generates a sequence of polygon vertices outlining the object. Due to its flexibility, it is capable of modeling the bounding box detection and polygon prediction tasks in a unified way.
    \item We introduce a novel framework for RIS and REC, called PolyFormer, which formulates them as a sequence-to-sequence prediction problem. Due to its flexibility, it can naturally fuse multi-modal features together as input and generate a sequence of polygon vertices and bounding box corner points.
    \vspace{-2mm}
    %\item We propose a regression-based decoder tailored for our framework. It predicts the continuous 2D vertex coordinates without quantization. We show that the proposed regression-based decoder  considerably improves coordinate prediction accuracy, compared with the standard classification-based decoder.
    \item We propose a regression-based decoder for accurate coordinate prediction in this seq2seq framework, which outputs continuous 2D coordinates directly without quantization error. To the best of our knowledge, this is the first work formulating geometric localization as a regression task in seq2seq framework instead of classification as in \cite{chen2021pix2seq,chen2022unified,wang2022unifying,lu2022unifiedio}.
    \vspace{-2mm}
    %\item For the first time, we show that a method with polygonal outputs surpasses mask-based methods by a large margin across all three main referring image segmentation benchmarks. We further demonstrate the effectiveness of PolyFormer by directly applying it for referring video segmentation, which achieves competitive performance compared with more complex methods specifically designed for video. 
    \item For the first time, we show that the polygon-based method surpasses mask-based ones across all three main referring image segmentation benchmarks, and it can also generalize well to unseen scenarios, including video and synthetic data. 
\end{itemize}

\begin{figure*}[t]
  \centering
\includegraphics[width=0.98\textwidth]{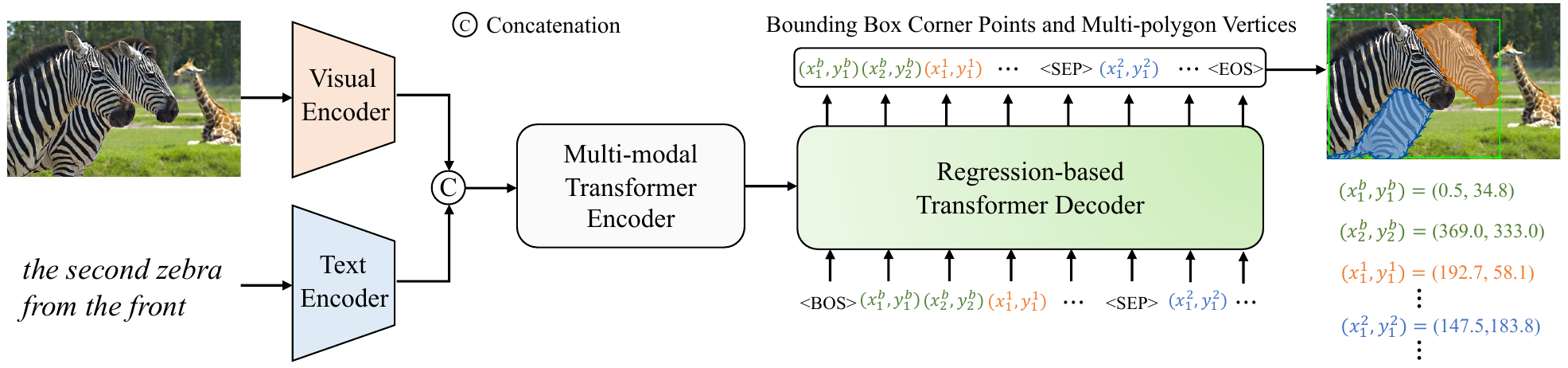}
  \caption{
%   Overall pipeline of the proposed PolyFormer. 
  Overview of our PolyFormer architecture.
  %It consists of four parts: a visual and text encoder, a multi-modal encoder and a regression-based decoder. 
%   The model takes an image and its corresponding language description as input, and generates the continuous 2D coordinates of the referred object bounding box and polygons in an autoregressive way.
 The model takes an image and its corresponding language expression as input, and outputs the floating-point 2D coordinates of the bounding box and polygons in an autoregressive way. 
%  \zc{is ``what region does the text...'' required? In fig. 1, not such prompt. and our method in fact doesn't need this prompt.}
% The same colors represent the coordinates belong to the same polygon or box.
  }
  \label{fig:polyformer_overview}
\end{figure*}

\vspace{-2.5mm}
\section{Related Work}
\paragraph{Referring Image Segmentation}
(RIS)~\cite{hu2016segmentation} aims to provide pixel-level localization of a target object in an image described by a referring expression. 
\iffalse
A typical pipeline involves two steps: (1) vision and language feature extraction, and (2) multi-modal feature fusion
\zc{nothing after fusion? maybe what want say is ``The previous work mainly focuses on two aspects: (1)... (2)...''}. 
\fi 
The previous works mainly focus on two aspects: (1) vision and language feature extraction, and (2) multi-modal feature fusion.
For feature extraction, there has been a rich line of work, including the use of CNNs~\cite{hu2016segmentation, liu2017recurrent,  yu2018mattnet, ye2019cross, chen2019referring, chen2019see, hu2020bi, huang2020referring, hui2020linguistic}, recurrent neural networks~\cite{schmidhuber1997lstm, chen2019referring, yu2018mattnet, cho2014gru, luo2020multi, feng2021encoder}, and transformer models~\cite{kim2022restr,yang2022lavt, li2021referring}. 
The efforts on feature fusion have explored feature concatenation~\cite{hu2016segmentation, liu2017recurrent}, attention mechanisms~\cite{shi2018key, chen2019see, ye2019cross, hu2020bi}, and multi-modal transformers~\cite{ding2021vlt, wang2022cris, li2021referring, kim2022restr}. The method most related to ours is SeqTR~\cite{zhu2022seqtr}, which also adopts a transformer model for generating the polygon vertices sequentially.
However, SeqTR can only produce a single polygon of 18 vertices with coarse segmentation masks, failing to outline objects with complex shapes and occlusion.

\vspace{-1mm}
\paragraph{Referring Expression Comprehension}
(REC) predicts a bounding box that tightly encompasses the target object in an image corresponding to a referring expression.
Existing works include two-staged methods~\cite{hu2017modeling, zhang2018grounding, zhuang2018parallel, hong2019learning} that are based on region proposal ranking, and one-stage methods~\cite{kamath2021mdetr, li2021referring, zhu2022seqtr, liao2020real, yang2019fast,cai2022x} that directly predict the target bounding box. Several papers ~\cite{li2021referring, zhu2022seqtr, luo2020multi} explore multi-task learning of REC and RIS since they are two closely related tasks.
However, MCN~\cite{luo2020multi} and RefTR~\cite{li2021referring} require task-specific heads. 
% Although SeqTR~\cite{zhu2022seqtr} casts the visual grounding tasks \zc{REC? use consistent naming} as a point prediction problem and designs a universal network \zc{so this is similar to what we do?}, 
Although SeqTR~\cite{zhu2022seqtr} casts the two tasks as a point prediction problem in a unified framework, 
it shows that multi-task supervision degenerates the performance compared with the single-task variant. In contrast, our PolyFormer achieves improved performance via multi-task learning of RIS and REC.

\vspace{-3mm}
\noindent \paragraph{Sequence-to-Sequence (seq2seq) Modeling} has achieved a lot of successes in natural language processing (NLP)~\cite{seq2seq, seq2seq2,radford2019language,raffel2020exploring,brown2020language}. 
Sutskever \etal~\cite{seq2seq} propose a pioneering seq2seq model based on LSTM~\cite{schmidhuber1997lstm} for machine translation.
Raffel \etal~\cite{raffel2020exploring} develop the T5 model to unify various tasks including translation, question answering and classification in a text-to-text framework.~\cite{brown2020language} further shows that scaling up language models significantly improves few-shot performance. Inspired by these successes in NLP,
recent endeavors in computer vision and vision-language also start to explore seq2seq modeling for various tasks~\cite{chen2021pix2seq, chen2022unified,yang2022unitab, wang2022unifying,lu2022unifiedio, zhu2022seqtr}. However, they cast geometric localization tasks as a classification problem, \ie, quantizing coordinates into discrete bins and predicting the coordinates as one of the bins. This enables them to unify all tasks into a simple unified seq2seq framework, but neglects the differences among tasks. In PolyFormer, geometric localization is formulated as a more suitable regression task that predicts continuous coordinates without quantization.
% directly without quantization.
%Inspired by these successes in NLP, recent endeavors in computer vision and vision-language also start to explore this simple and unified seq2seq modeling ~\cite{chen2021pix2seq, chen2022unified,yang2022unitab, wang2022unifying,lu2022unifiedio, zhu2022seqtr}. However, they cast geometric localization tasks as classification, \ie, quantizing coordinates into discrete bins and predicting the hypothesis to one of bins. This enables to unify all tasks into the simple seq2seq framework, but it neglects the differences among various tasks. In PolyFormer, the geometric localization is formulated as the more suitable regression task to predict continuous coordinates directly without quantization.
%ignore the differences between classification and localization tasks, and cast all localization existing methods adopt a serialization and quantization scheme to convert the bounding box or segmentation mask into a sequence of discrete tokens, which are then predicted via a softmax classification layer.
%To the best of our knowledge, this is the first paper on designing a regression-based decoder to directly output the continuous coordinates.

% \vspace{-3.25mm}
\vspace{-2.5mm}
\paragraph{Contour-based Instance Segmentation} 
aims to segment instances by predicting the contour.
~\cite{castrejon2017annotating} labels object instances with polygons via a recurrent neural network.
PolarMask~\cite{xie2020polarmask} models instance masks in polar coordinates, and converts instance segmentation to instance center classification and dense distance regression tasks. Deep Snake~\cite{DBLP:conf/cvpr/PengJPLBZ20} extends the classic snake algorithm~\cite{snake} and uses a neural network to iteratively deform an initial
contour to match the object boundary.
PolyTransform~\cite{polytransform} exploits a segmentation network to first generate instance masks to initialize polygons, which are then fed into a deformation network to better fit the object boundary. BoundaryFormer~\cite{boundaryformer} introduces a point-based transformer with mask supervision via a differentiable rasterizer. However, these papers are limited in how they handle fragmented objects.

% \section{The PolyFormer Framework}
% \zc{using just PolyFormer as the section title looks better?}
\section{PolyFormer}

% \subsection{Architecture Overview}

% \cref{fig:polyformer_overview} illustrates the framework \zc{architecture in section title? pipeline in the figure? should be consistent a little bit} of our Polygon Transformer (PolyFormer). Instead of predicting dense segmentation masks, PolyFormer sequentially produces the corner points of the bounding box and the vertices of polygons outlining the object.
% Specifically, we first use a visual encoder and a text encoder to extract image and text features, respectively, which are then projected into a shared embedding space. Next, we concatenate the image and text features, and feed them into a multi-modal transformer encoder. Finally, a regression-based transformer decoder takes the encoded features and outputs the continuous floating-point bounding box points and polygon vertices in an autoregressive way. The segmentation mask is generated as the region compassed by the polygons. 

\subsection{Architecture Overview}
\cref{fig:polyformer_overview} gives an overview of PolyFormer architecture. Instead of predicting dense segmentation masks, PolyFormer sequentially produces the corner points of the bounding box and vertices of polygons outlining the object.
Specifically, we first use a visual encoder and a text encoder to extract image and text features, respectively, which are then projected into a shared embedding space. Next, we concatenate the image and text features, and feed them into a multi-modal transformer encoder. Finally, a regression-based transformer decoder takes the encoded features and outputs the continuous floating-point bounding box corner points and polygon vertices in an autoregressive way. The segmentation mask is generated as the region encompassed by the polygons.

% \subsection{Target Polygon Sequence Construction}
\subsection{Target Sequence Construction}
\label{subsed:polygon construction}

We first describe how to represent ground-truth bounding box and polygon sequences.

\paragraph{Polygon Representation.}

\begin{figure}[t]
  \centering
  \includegraphics[width=0.5\textwidth]{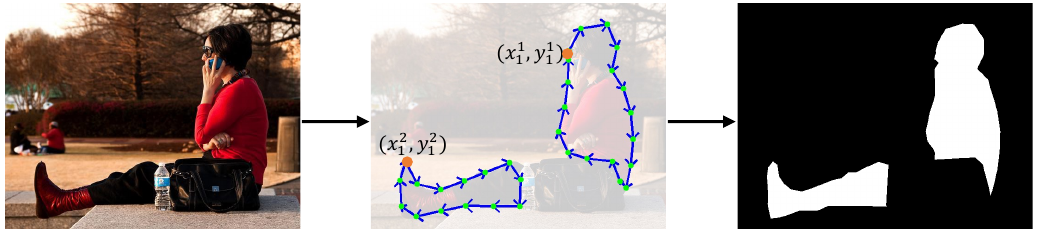}
  \caption{
%   Polygon sequence representation. The vertices in a polygon are sorted in a clock-wise order, where the starting points (orange dots) are the vertices that are closest to the image origin. Once the polygons are obtained, we can get the binary segmentation mask easily.
Illustration of polygon sequence representation. The vertices in a polygon are sorted in clockwise order, where the starting points (orange dots) are the vertices that are closest to the image origin. The segmentation mask is generated as the region encompassed by the polygons. 
%The binary segmentation mask is generated by Once the polygons are obtained, it is straightforward to get .
%   accordingly.
  }
  \label{fig:seq}
  \vspace{-2mm}
\end{figure}

A segmentation mask is described using one or more polygons outlining the referred object.
We parameterize a polygon as a sequence of 2D vertices $\{(x_i,y_i)\}_{i=1}^{K}$
, $(x_i,y_i)\in \mathbb{R}^2$ in the clock-wise order.
We choose the vertex that is closest to the top left corner of the image as the starting point of the sequence (see \cref{fig:seq}).

\paragraph{Vertex and Special Tokens.}
For each vertex coordinate $x$ or $y$, previous works~\cite{chen2021pix2seq,chen2022unified,wang2022unifying,lu2022unifiedio,yang2022unitab,zhu2022seqtr} uniformly quantize it into an integer between $[1, B]$, where $B\in \mathbb{N}$ is the number of bins of the coordinate codebook. In contrast, we maintain the continuous floating-point value of the original $x$ or $y$ coordinate without any quantization. 
To represent multiple polygons, we introduce a separator token $\texttt{<SEP>}$ between two polygons. Polygons from the same object are ordered based on the distance between their starting points and the image origin.
Finally, we use $\texttt{<BOS>}$ and $\texttt{<EOS>}$ tokens to indicate the beginning and end of the sequence.

\paragraph{Unified Sequence with Bounding Box.}
A bounding box is represented by two corner points, \ie, top-left $(x_{1}^b,y_{1}^b)$ and bottom-right $(x_{2}^b,y_{2}^b)$. The coordinates of the bounding box and multiple polygons can be concatenated together into a single long sequence as follows:
\begin{align}
[&\texttt{<BOS>},(x_{1}^b,y_{1}^b),(x_{2}^b,y_{2}^b), (x_1^1,y_1^1), \nonumber
\\ \nonumber
&(x_2^1,y_2^1),...,\texttt{<SEP>},(x_1^n,y_1^n),...,\texttt{<EOS>}], \nonumber
\end{align}
where $(x_1^n,y_1^n)$ is the starting vertex of the $n^{th}$ polygon. In general, the bounding box corner points and polygon vertices are regarded as the coordinate tokens $\texttt{<COO>}$.

\subsection{Image and Text Feature Extraction}
As illustrated in \cref{fig:polyformer_overview}, the input of our framework consists of an image $I$ and a referring expression $T$.

\noindent \paragraph{Image Encoder.}
For an input image $I\in \mathbb{R}^{H\times W\times 3}$, we use a Swin transformer~\cite{liu2021swin} to extract the feature from the 4-th stage as visual representation $F_v\in \mathbb{R}^{\frac{H}{32}\times \frac{W}{32}\times C_v}$.

\noindent \paragraph{Text Encoder.}
Given the language description $T\in \mathbb{R}^{L}$ with $L$ words, we use the language embedding model from BERT~\cite{devlin2018bert} to extract the word feature $F_l\in \mathbb{R}^{L\times C_l}$.

\noindent \paragraph{Multi-modal Transformer Encoder.}
To fuse the image and textual features, we flatten  $F_v$ into a sequence of visual features $F'_v\in \mathbb{R}^{(\frac{H}{32}\cdot \frac{W}{32})\times C_v}$ and project $F'_v$ and $F_l$ into the same embedding space with a fully-connected layer: 
\begin{equation}
F'_v = F'_v W_v + b_v,\ F'_l = F_lW_l + b_l,
\end{equation}
where $W_v$ and $W_l$ are learnable matrices to transform the visual and textual representations into the same feature dimension, $b_v$ and $b_l$ are the bias vectors. The projected image and text features are then concatenated: $F_M = [F'_v, F'_l].$

The multi-modal encoder is composed of $N$ transformer layers, where each layer consists of a multi-head self-attention layer, a layer normalization and a feed-forward network. It takes the concatenated feature $F_M$ and generates the multi-modal feature $F_M^N$ progressively.

To preserve position information, absolute positional encodings~\cite{ke2020rethinking} are added to the image and text features. In addition, we add 1D~\cite{raffel2020exploring} and 2D~\cite{dai2021coatnet,wang2021simvlm} relative position bias to image and text features, respectively.

\begin{figure}[t]
  \centering
  \includegraphics[width=0.45\textwidth]{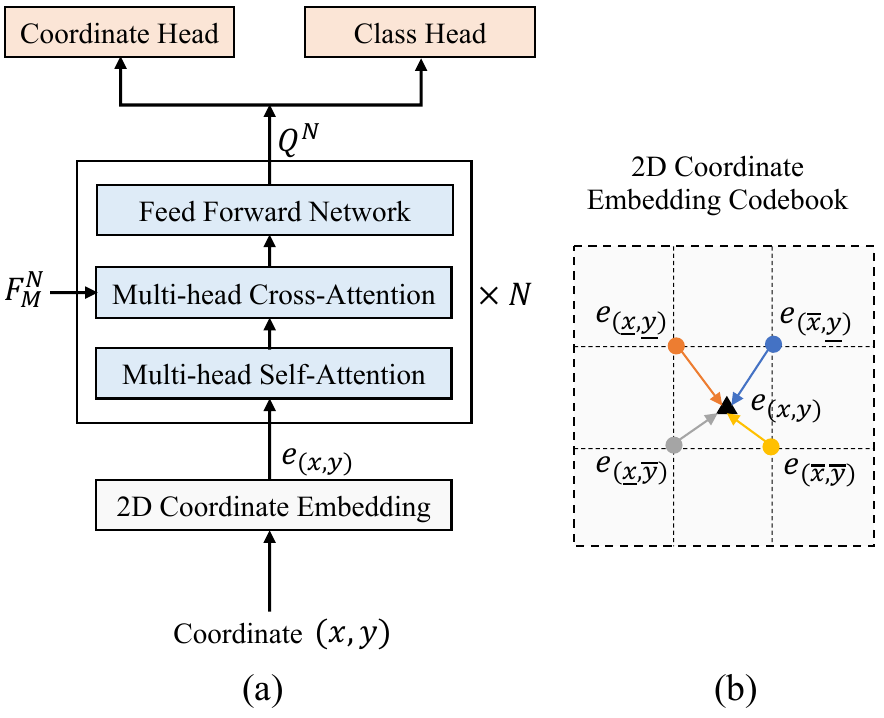}
  \caption{The architecture of the regression-based transformer decoder (a). 
%   The generation process of 2D coordinate embedding is illustrated in (b).
  The 2D coordinate embedding is obtained by bilinear interpolation from the nearby grid points, as illustrated in (b). %No quantization is performed on any coordinates.
  }
 
  \label{fig:decoder}
  \vspace{-2mm}
\end{figure}

\subsection{Regression-based Transformer Decoder}
Previous visual seq2seq methods~\cite{chen2021pix2seq,chen2022unified,yang2022unitab,zhu2022seqtr,lu2022unifiedio} quantize a continuous coordinate $x$ into a discrete bin $[x]$, introducing an inevitable quantization error $|x-[x]|$.  They formulate coordinate localization as a classification problem to predict one of $[x]$s, which is suboptimal for geometric localization. To address this issue, we propose a regression-based decoder that does not use quantization, and instead predicts the \textit{continuous} coordinate values directly (\textit{i.e.}, we use $x$ instead of $[x]$), as shown in \cref{fig:decoder}.

\paragraph{2D Coordinate Embedding.} 
In~\cite{chen2021pix2seq,chen2022unified,yang2022unitab,zhu2022seqtr,lu2022unifiedio}, the coordinate codebook is in 1D space, $\mathcal{D}\in \mathbb{R}^{B \times C_e}$,
where $B$ is the number of bins, $C_e$ is the embedding dimension. The embedding of $x$ is
obtained by indexing the codebook, \ie, $\mathcal{D}([x])$. 
% For a more accurate representation, 
\iffalse
For a more accurate coordinate representation,
we build a 2D coordinate codebook, $\mathcal{D}\in \mathbb{R}^{B_{H} \times B_{W} \times C_e}$, where $B_{H}$ and $B_{W}$ are the numbers of bins along the height and width dimensions, respectively.
% The 2D coordinate codebook provides a more accurate coordinate embedding for any floating-point coordinate $(x,y)\in \mathbb{R}^2$.
The 2D coordinate codebook provides a precise coordinate embedding for any floating-point coordinate $(x,y)\in \mathbb{R}^2$.
\fi
To better capture the geometric relationship between $x$ and $y$ and have a more accurate coordinate representation,
we build a 2D coordinate codebook, $\mathcal{D}\in \mathbb{R}^{B_{H} \times B_{W} \times C_e}$, where $B_{H}$ and $B_{W}$ are the numbers of bins along height and width dimensions.
With this 2D coordinate codebook, we can obtain the precise coordinate embedding for any floating-point coordinate $(x,y)\in \mathbb{R}^2$.
First, the floor and ceiling operations are applied on $(x,y)$ to generate four discrete bins: $(\underline{x},\underline{y}), (\bar{x},\underline{y}), (\underline{x},\bar{y}), (\bar{x},\bar{y})\in \mathbb{N}^2$, and the corresponding embeddings can be indexed from the 2D codebook, \eg, $e_{(\underline{x},\underline{y})}=\mathcal{D}(\underline{x},\underline{y})$. Finally, we get the accurate coordinate embedding $e_{(x,y)}$ by bilinear interpolation, as in  \cite{he2017mask}:
% \vspace{-2mm}
\begin{equation}
\begin{aligned}
e_{(x,y)} = & (\bar{x}-x)(\bar{y}-y)\cdot e_{(\underline{x},\underline{y})} + (x - \underline{x})(\bar{y}-y)\cdot e_{(\bar{x},\underline{y})} + \\
& (\bar{x}-x)(y - \underline{y})\cdot e_{(\underline{x},\bar{y})} + (x-\underline{x})(y-\underline{y})\cdot e_{(\bar{x},\bar{y})}.
% ,x\in \mathbb{R},\underline{x} \&  \hat{x}\in \mathbb{I}
\end{aligned}
\end{equation}
\vspace{-7.5mm}
\paragraph{Transformer Decoder Layers.} To capture the relations between multi-modal feature $F_M^N$  and 2D coordinate embedding $e_{(x,y)}$, we introduce $N$ transformer decoder layers. Each transformer layer consists of a multi-head self-attention layer, a multi-head cross-attention layer and a feed-forward network. 

% \vspace{-4mm}
\paragraph{Prediction Heads.} Two lightweight heads are built on top of the last decoder layer output $Q^N$ to generate final predictions. The class head is a linear layer that outputs the token types, indicating whether the current output is a coordinate token ($\texttt{<COO>}$), separator token ($\texttt{<SEP>}$) or an end-of-sequence token ($\texttt{<EOS>}$):
\begin{align}
\hat{p} &= W_cQ^N+b_c,
\end{align}
where $W_c$ and $b_c$ are parameters of the linear layer.

The coordinate head is a 3-layer feed-forward network (FFN) with ReLU activation except for the last layer. It predicts the 2D coordinates of the referred object bounding box corner points and polygon vertices:
\begin{align}
(\hat{x},\hat{y}) &= Sigmoid(FFN(Q^N)).
\end{align}

\subsection{Training}
\label{sec:training}

\begin{figure}[t]
  \centering
  \includegraphics[width=0.5\textwidth]{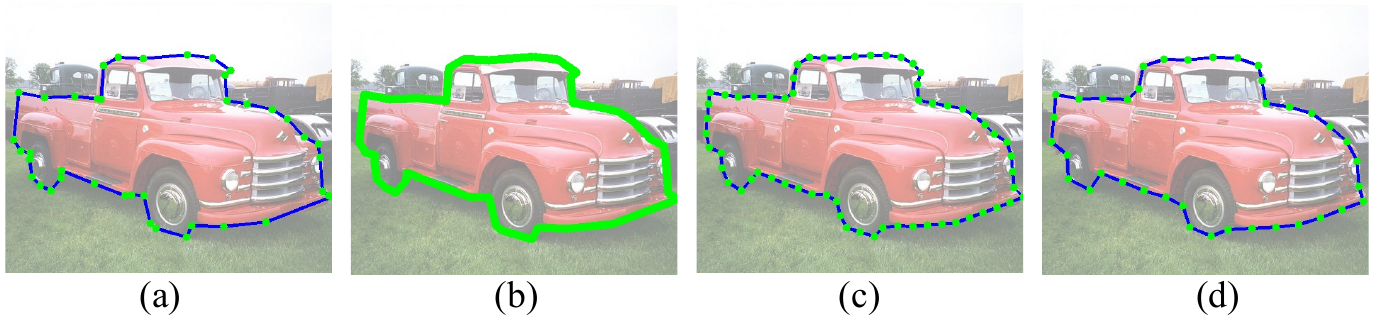}
  \caption{
%   Polygon augmentation. 
%   Polygons at different levels of granularity (c) - (d) are generated by dense interpolation (b) based on the ground-truth polygon (a).
%Illustration of polygon augmentation. Sparse polygons (c)-(d) are uniformly down-sampled from densely interpolated object contour (b) based on the ground-truth polygon (a).
Illustration of polygon augmentation. 
%Dense contour (b) are obtained from the ground-truth polygon (a) by interpolation. Augmented polygons (c)-(d) are generated from (b) via uniform downsampling.
% Given (b), augmented polygons (c)-(d) are randomly generated via uniform down-sampling.
Polygons at different levels of granularity (c) - (d) are sampled from dense contour (b) that is interpolated from the original polygon (a).
  }
  \label{fig:aug}
  \vspace{-3mm}
\end{figure}

\paragraph{Polygon Augmentation.}
A polygon is a sparse representation of the dense object contour. Given a dense contour, the generation of sparse polygons is usually not unique. Given this property,
we introduce a simple yet effective augmentation technique to increase polygon diversity. As illustrated in \cref{fig:aug}, the dense contour is first interpolated from the original polygon. Then, uniform down-sampling is applied with an interval randomly sampled from a fixed range to generate sparse polygons. This creates diverse polygons at different levels of granularity, and prevents the model from being overfitted to a fixed polygon representation.

\input{tables/res_v3}

\paragraph{Objective.}
Given an image, a referring expression and preceding tokens, the model is trained to predict the next token and its type:
\begin{equation}
\begin{aligned}
L_t &= \lambda_{t} L_{coo}((x_t,y_t), (\hat x_t,\hat y_t)|I, T, (x_i,y_i)_{i=1:t-1}) \\ 
& \cdot \mathbb{I}[p_t==\texttt{<COO>}] 
 + \lambda_{cls} L_{cls}(p_t, \hat p_t|I, T, p_{1:t-1}), \\
\end{aligned}
\label{eq:loss}
\end{equation}
where
\begin{equation*}
\lambda_t=\left\{
\begin{aligned}
\lambda_{box}, &~~~~t \leq 2,\\
\lambda_{poly},&~~~~otherwise, \\
\end{aligned}
\right.
\end{equation*}

\noindent $L_{coo}$ is the $L_1$ regression loss, $L_{cls}$ is the label smoothed cross-entropy loss, and $\mathbb{I}[\cdot]$ is the indicator function. The regression loss is only computed for the coordinate tokens, where $\lambda_{box}$ and $\lambda_{poly}$ are the corresponding token weights. The total loss is the sum of $L_t$ over all tokens in a sequence.

\paragraph{Inference.}
During inference, we start the generation by inputting the \texttt{<BOS>} token. First, we get the token type from the class head. If it is a coordinate token, we will obtain the 2D coordinate prediction from the coordinate head conditioned on the preceding predictions; if it is a separator token, it indicates the end of the preceding polygon, so the separator token will be added to the output sequence. This sequential prediction will stop once \texttt{<EOS>} is output. 
In the generated sequence, the first two tokens are bounding box coordinates and the rest are polygon vertices. 
The final segmentation mask is obtained from the polygon predictions.

\vspace{-1mm}
\section{Experimental Results}
\vspace{-1mm}
\subsection{Datasets and Metrics}
\vspace{-3mm}
\noindent \paragraph{Datasets.}
The experiments are conducted on four major benchmarks for RIS and REC: RefCOCO \cite{yu2016modeling}, RefCOCO+ \cite{yu2016modeling}, RefCOCOg \cite{mao2016generation, nagaraja2016modeling}, and ReferIt~\cite{referit}. RefCOCO has 142,209 annotated expressions for 50,000 objects in 19,994 images, and RefCOCO+ consists of 141,564 expressions for 49,856 objects in 19,992 images. Compared with RefCOCO, location words are absent from the referring expressions in RefCOCO+, which makes it more challenging. RefCOCOg consists of 85,474 referring expressions for 54,822 objects in 26,711 images. The referring expressions are collected on Amazon Mechanical Turk, and therefore the descriptions are longer and more complex (8.4 words on average \textit{vs.}~3.5 words of RefCOCO and RefCOCO+). ReferIt contains 130,364 expressions for 99,296 objects in 19,997 images collected from the SAIAPR-12 dataset~\cite{ESCALANTE2010419}. We use the UMD split for RefCOCOg~\cite{nagaraja2016modeling} and Berkeley split for ReferIt.  

\vspace{-1mm}
\noindent \paragraph{Evaluation Metrics.}
We use mean Intersection-over-Union (mIoU) as the evaluation metric for RIS. For a fair comparison, we also use overall Intersection-over-Union (oIoU) when comparing with papers that only report oIoU results.
Additionally, we also evaluate PolyFormer on the REC task as it is a unified framework for both RIS and REC tasks. We adopt the standard metric Precision@0.5~\cite{zhu2022seqtr, wang2022unifying}, where a prediction is considered correct if its Intersection-over-Union (IoU) with the ground-truth box is higher than 0.5.

\input{tables/rec_v2}
\vspace{-1mm}
\subsection{Implementation Details}
\noindent \textbf{Model Settings.}
In PolyFormer-B, we use Swin-B~\cite{liu2021swin} as the visual encoder and BERT-base~\cite{devlin2018bert} as the text encoder.
For the transformer encoder and decoder, we adopt 6 encoder layers and 6 decoder layers. 
To investigate the impact of different model scales, we develop a larger model, PolyFormer-L, that adopts Swin-L as the visual backbone with 12 transformer encoder and decoder layers.
% \zc{the number of coordinate bins is not introduced anywhere?}
% \vspace{-2mm}

\noindent \paragraph{Training Details.}
To leverage large-scale image-text dataset with grounded box annotations, we first pre-train PolyFormer on the REC task with the combination of Visual Genome~\cite{krishnavisualgenome}, RefCOCO~\cite{yu2016modeling}, RefCOCO+~\cite{yu2016modeling}, RefCOCOg~\cite{mao2016generation, nagaraja2016modeling}, and Flickr30k-entities~\cite{plummer2015flickr30k}. 
% Following~\cite{cai2022x}, we do multi-task finetuning  for both RIS and REC on a combined training set of RefCOCO, RefCOCO+, and RefCOCOg with all the validation and testing images removed. 
% During the multi-task fine-tuning stage, we train the model for both RIS and REC on a combined training dataset~\cite{cai2022x} of RefCOCO, RefCOCO+, and RefCOCOg with all validation and testing images removed.
During the multi-task fine-tuning stage, for RefCOCO, RefCOCO+, and RefCOCOg datasets, we train the model for both RIS and REC on a combined training dataset~\cite{cai2022x} with all validation and testing images removed; for ReferIt~\cite{referit} dataset, only the ReferIt training set is used. We use the AdamW optimizer~\cite{loshchilov2017decoupled} with $(\beta_1, \beta_2)=(0.9, 0.999)$ and $\epsilon=1\times 10^{-8}$. The initial learning rate is $5\times 10^{-5}$ with polynomial learning rate decay. We train the model for 20 epochs during pre-training and 100 epochs for fine-tuning with a batch size of 160 and 128, respectively.
The coefficients for losses are set as $\lambda_{box}=0.1$, $\lambda_{poly}=1$ and $\lambda_{cls}=5\times 10^{-4}$. Images are resized to $512 \times 512$, and polygon augmentation is applied with a probability of 50\%. The 2D coordinate embedding codebook is constructed with $64\times64$ bins.

% \vspace{-7.2mm}
\vspace{-1mm}
\subsection{Main Results}
\vspace{-1mm}
\paragraph{Referring Image Segmentation.}
We compare PolyFormer with the state-of-the-art methods in Table~\ref{tab:res}. It can be observed that PolyFormer models outperform previous methods on each split of the three datasets under all metrics by a clear margin. First, on the RefCOCO dataset, compared with the recent LAVT~\cite{yang2022lavt}, PolyFormer-B achieves better results with absolute mIoU gains of 1.5\%, 0.2\%, and 2.28\% on the three splits. Second, on the more challenging RefCOCO+ dataset, PolyFormer-B significantly outperforms the previous state-of-the-art RefTr~\cite{li2021referring} by absolute mIoU margins of 3.9\%, 3.93\%, 5.24\% on the validation, test A and test B sets, respectively. Third, on the most challenging RefCOCOg dataset where language expressions are longer and more complex, PolyFormer-B achieves notable performance improvements of 2.73\% and 2.49\% mIoU points on the validation and test sets compared with the second-best method RefTr~\cite{li2021referring}. Using the stronger Swin-L backbone and a larger encoder and decoder, PolyFormer-L achieves consistent performance gains of around 1$\sim$2 absolute points over PolyFormer-B across all datasets. These results demonstrate the superiority of our polygon-based method PolyFormer over the previous mask-based methods.  
% \zc{use PolyFormer-B and PolyFormer-L?}
% \zc{not Swin-B but PolyFormer-B} 
\vspace{-2mm}

\input{tables/rvs}

\paragraph{Referring Expression Comprehension.}
We further evaluate PolyFormer on the REC datasets and compare the performance with other state-of-the-art methods in Table~\ref{tab:rec}. OFA~\cite{wang2022unifying} is a foundation model which provides a unified interface for a wide range of tasks from different modalities. Compared with OFA-L, PolyFormer-L achieves better results on RefCOCO and comparable results on the remaining two datasets. Note that OFA utilizes 19M image-text data for pre-training, while PolyFormer only uses 6M image-text pairs. Another recent paper SeqTR~\cite{zhu2022seqtr} also proposes a seq2seq model for multi-task learning of RIS and REC, but they observe performance degradation for each task when trained jointly. PolyFormer-B significantly outperforms SeqTR by absolute margins of 2.73\%, 5.04\% and 1.77\% on the three validation sets. These results illustrate the effectiveness of our unified framework.

\vspace{-1mm}
\paragraph{Zero-shot Transfer to Referring Video Object Segmentation.}
To further test its generalization ability, we evaluate PolyFormer on the referring video object segmentation dataset Ref-DAVIS17~\cite{khoreva2018video} in a zero-shot manner. We simply view the video data as a sequence of images, and apply PolyFormer to the video frame-by-frame. As shown in Table~\ref{tab:rvs-davis}, PolyFormer-L achieves 61.5\% $\mathcal{J}\&\mathcal{F}$ without any finetuning on the video data. It is even better than the state-of-the-art ReferFormer~\cite{wu2022referformer}, which is fully trained on referring video segmentation data.

\vspace{-1mm}
\subsection{Ablation Studies}
In this section, we perform extensive ablation studies on the RefCOCO, RefCOCO+ and RefCOCOg validation sets to study the effects of core components of PolyFormer. 
% All models are based on Swin-B visual backbone with 6 encoder and decoder layers unless stated otherwise 
All ablation experiments are performed on PolyFormer-B.
% \zc{no experiments on Large model, right? if so, remove ``unless stated otherwise''} \jl{removed}

\input{tables/regression}

\paragraph{Coordinate Classification \textit{vs.} Regression.}
We implement a classification-based model for coordinate prediction following~\cite{chen2021pix2seq, yang2022unitab, zhu2022seqtr}. Specifically, the coordinates are quantized into discrete bins and a classifier is adopted to output the discrete tokens. As shown in Table~\ref{tab:decoder}, the regression-based model consistently outperforms the classification-based model on all datasets, \textit{e.g.}, +1.86 for RIS and +2.38 for REC on the RefCOCO+ dataset. This shows that the regression-based model is a better choice for geometric localization tasks than the classification-based counterpart.

\begin{table}[t]
\centering
\scalebox{0.88}{
\setlength{\tabcolsep}{0.6mm}{\begin{tabular}{c|c|c|c|c|c|c}
\toprule
{Order}       & {Aug}       & {Multi-task}       & {\texttt{<SEP>} }        & RefCOCO & RefCOCO+ & RefCOCOg  \\
\midrule
    \xmark       &    \xmark        &       \xmark     &     \xmark        & 55.92    & 51.64 & 50.65      \\
\checkmark &     \xmark       &    \xmark        &    \xmark         &   68.35   &   63.28 & 62.18    \\
\checkmark & \checkmark &   \xmark         &       \xmark      &    72.07  &    66.68 & 65.20   \\
\checkmark & \checkmark & \checkmark &  \xmark           &   75.14   &     69.86 & 68.70  \\
\checkmark & \checkmark & \checkmark &  \checkmark & 75.96     & 70.65 & 69.36 \\
\bottomrule
\end{tabular}}}
\caption{
% Ablation study on different components of target sequence construction.
Ablation study on target sequence construction.
% Impact of different components
% \zc{put tables/figure on the top is super simple, just have [t], please don't change to [h] or something else anymore.}
}
\vspace{-2mm}
\label{tab:ablation-data}
\end{table}

\input{figs/attn_v3}
\input{figs/vis}

\paragraph{Component Analysis of Target Sequence Construction.}
We study the effects of several components in constructing the target sequence, including (1) polygon ordering, where the polygon vertices are ordered clock-wise and the starting point is set as the vertex closest to the image origin; (2) data augmentation, where we generate polygons  at different levels of granularity on-the-fly during training; (3) multi-task learning of RIS and REC; and (4) separator token, where we add \texttt{<SEP>} token to handle multi-polygon cases.
The results are shown in Table~\ref{tab:ablation-data}. Using randomly ordered polygons only achieves 55.92\% mIoU on RefCOCO. Polygon ordering is essential and leads to a substantial improvement of 12.43\%. Both polygon augmentation and multi-task learning are beneficial, with gains of 3.72\% and 3.07\%, respectively. The separator token \texttt{<SEP>} brings a 0.82\% increase of mIoU. This small gain is reasonable considering only a small number of samples have multiple polygons. We observe similar trends in the remaining two datasets.

\subsection{Visualization Results}
\paragraph{Cross-attention Map.}
When generating the vertex tokens, the regression-based decoder computes the self-attention over the preceding tokens and cross-attention over the multi-modal feature. Here we visualize the cross-attention map (averaged over all layers and heads) when the model predicts a new token.~\cref{fig:attn} shows the cross-attention maps at different steps of the polygon generation. We observe that the cross-attention map concentrates on the object referred to by the sentence, and moves around the object boundary during the polygon generation process. 

% \vspace{-15mm}
\paragraph{Prediction Visualization.}
We show the visualization results of PolyFormer on the RefCOCOg test set in~\cref{fig:visualization} (a)-(d). It can be seen that PolyFormer is able to segment the referred object in challenging scenarios, \textit{e.g.}, instances with occlusion and complex shapes, and instances that are partially displayed or require complex language understanding. We also show the results on images generated by Stable Diffusion~\cite{rombach2021highresolution} in~\cref{fig:visualization} (e)-(g). PolyFormer demonstrates good generalization ability on synthetic images and text descriptions that have never been seen during training. In contrast, the state-of-the-art LAVT \cite{yang2022lavt} and SeqTR \cite{zhu2022seqtr} fail to generate satisfactory results. More visualization results are provided in the supplementary material.

\vspace{-4mm}
\section{Conclusion}
In this work, we propose PolyFormer, a simple and unified framework for referring image segmentation and referring expression comprehension. It is a sequence-to-sequence framework that can naturally fuse multi-modal features as the input sequence and multi-task predictions as the output sequence. 
%Thus the same model architecture and training objective can be adopted for these two different tasks. 
Moreover, we design a novel regression-based decoder to generate continuous 2D coordinates without quantization errors. PolyFormer achieves competitive results for RIS and REC and shows good generalization to unseen scenarios. We believe this simple framework can be extended to other tasks beyond RIS and REC. 
% \zc{any discussion on limitations and negatively social impact required by CVPR this year?}
% \iffalse
% PolyFormer consistently outperforms previous state-of-the-art methods on three referring image segmentation and referring expression comprehension benchmarks, demonstrating its effectiveness and superiority. 
% \fi
%PolyFormer outperforms (or is on-par-with) previous state-of-the-art methods on three referring image segmentation and referring expression comprehension benchmarks, demonstrating its effectiveness and superiority.
%We further test our model on the referring video object segmentation dataset in a zero-shot manner, and it also shows the state-of-the-art performance.

% \paragraph{Limitations and Broader Impacts.}
% % \subsection{Limitations and Broader Impacts}
% The training of PolyFormer requires accurate bounding box and polygon annotations. How to reduce such dependence and utilize weakly-supervised data for region-level image understanding needs further exploration. For the data and model, we need to further understand the broader impacts including but not limited to fairness, social bias and potential misuse.

% \input{appendix}

%\newpage
%%%%%%%%% REFERENCES
{\small
\bibliographystyle{ieee_fullname}
\bibliography{ref_new}
}

% \newpage
\clearpage
\appendix
\section*{Appendix}

\input{appendix}

\end{document}

%% file: tables/res_v3.tex
\begin{table*}[t]
   \centering
   \scalebox{0.941}{\setlength{\tabcolsep}{2.0mm}{\begin{tabular}{l|l|l|l|c|c|c|c|c|c|c|c|c}
      \toprule[1pt]
      %\hline
      \multicolumn{2}{c|}{\multirow{2}{*}{Method}} & Visual &
      Text & 
      \multicolumn{3}{c|}{RefCOCO}  & \multicolumn{3}{c|}{RefCOCO+} & \multicolumn{2}{c|}{RefCOCOg} & ReferIt\\
      \cline{5-13}
                                  \multicolumn{2}{c|}{}  & Backbone & Encoder & val   & test A & test B & val & test A & test B & val & test & test \\
      \hline
      \multirow{13}{*}{\rotatebox{90}{oIoU}} & STEP~\cite{chen2019see}       & RN101 & Bi-LSTM & 60.04 & 63.46 & 57.97 & 48.19 & 52.33 & 40.41 & -     & -   & 64.13  \\
      & BRINet~\cite{hu2020bi}    & RN101 & LSTM & 60.98 & 62.99 & 59.21 & 48.17 & 52.32 & 42.11 & -     & -   & 63.11  \\
      & CMPC~\cite{huang2020referring}     & RN101 & LSTM & 61.36 & 64.53 & 59.64 & 49.56 & 53.44 & 43.23 & -     & -  &  65.53  \\
      & LSCM~\cite{hui2020linguistic} & RN101 & LSTM & 61.47 & 64.99 & 59.55 & 49.34 & 53.12 & 43.50 & -     & -   & 66.57   \\
      & CMPC+~\cite{liu2021cross} &  RN101  & LSTM & 62.47 & 65.08 & 60.82 & 50.25 & 54.04 & 43.47 & -     & -   & 65.58  \\
      & MCN~\cite{luo2020multi}       & DN53 & Bi-GRU & 62.44 & 64.20 & 59.71 & 50.62 & 54.99 & 44.69 & 49.22 & 49.40 & -   \\
      & EFN~\cite{feng2021encoder}                & WRN101 & Bi-GRU & 62.76 & 65.69 & 59.67 & 51.50 & 55.24 & 43.01 & - & - & 66.70  \\
      & BUSNet~\cite{yang2021bottom}          & RN101 & Self-Att & 63.27 & 66.41 & 61.39 & 51.76 & 56.87 & 44.13 & - & - & - \\
      & CGAN~\cite{luo2020cascade}    & DN53 & Bi-GRU & 64.86 & 68.04 & 62.07 & 51.03 & 55.51 & 44.06 & 51.01 & 51.69 & - \\
      & LTS~\cite{jing2021locate}   & DN53 & Bi-GRU & 65.43 & 67.76 & 63.08 & 54.21 & 58.32 & 48.02 & 54.40 & 54.25 & - \\
      & ReSTR~\cite{kim2022restr}  & ViT-B & Transformer & 67.22 & 69.30 & 64.45 & 55.78 & 60.44 & 48.27 & - & - & 70.18\\ 
      \cline{2-13}
      & PolyFormer-B & Swin-B & BERT-base & 74.82 & 76.64 & 71.06 & 67.64 & 72.89 & 59.33 & 67.76 & 69.05 & 71.91 \\
      & \textbf{PolyFormer-L}  & Swin-L & BERT-base & \textbf{75.96} & \textbf{78.29} & \textbf{73.25} & \textbf{69.33} & \textbf{74.56} & \textbf{61.87} & \textbf{69.20} & \textbf{70.19} & \textbf{72.60}\\
      
      \hline
      
      \multirow{7}{*}{\rotatebox{90}{\textbf{mIoU}}} & VLT~\cite{ding2021vlt}    &  DN53 & Bi-GRU & 65.65 & 68.29 & 62.73 & 55.50 & 59.20 & 49.36 & 52.99 & 56.65 & - \\ % mIoU
      & CRIS~\cite{wang2022cris}  & RN101 & GPT-2 & 70.47 & 73.18 & 66.10 & 62.27 & 68.06 & 53.68 & 59.87 & 60.36 & - \\ %not sure if it's oIoU or mIoU
      & SeqTR~\cite{zhu2022seqtr} & DN53 & Bi-GRU & 71.70 & 73.31 & 69.82 & 63.04 & 66.73 & 58.97 & 64.69 & 65.74 & - \\
      & RefTr~\cite{li2021referring} & RN101 & BERT-base & 74.34 &76.77 & 70.87 & 66.75 & 70.58 & 59.40 & 66.63 & 67.39 & - \\
      & LAVT~\cite{yang2022lavt}  & Swin-B & BERT-base & {74.46} & {76.89} & {70.94} & {65.81} & {70.97} & {59.23} & {63.34} & {63.62} & - \\

      \cline{2-13}
       & PolyFormer-B & Swin-B & BERT-base & 75.96 & 77.09 & 73.22 & 70.65 & 74.51 & 64.64 & 69.36 & 69.88 & 65.98\\
      & \textbf{PolyFormer-L}  & Swin-L & BERT-base  & \textbf{76.94} & \textbf{78.49} & \textbf{74.83} & \textbf{72.15} & \textbf{75.71} & \textbf{66.73} & \textbf{71.15} & \textbf{71.17} & \textbf{67.22}\\
      \bottomrule[1pt]
      %\hline
   \end{tabular}}}
  \caption{Comparison with the state-of-the-art methods on three referring image segmentation benchmarks. 
%   \iffalse
  RN101 denotes ResNet-101~\cite{he2016deep}, WRN101 refers to Wide ResNet-101~\cite{zagoruyko2016wide}, and DN53 denotes Darknet-53~\cite{redmon2018yolov3}.
%   \fi
  }
   \label{tab:res}
%   We report the results of our method with various visual backbones, encoders and decoders. mIoU is utilized as the metric.}
\end{table*}

%% file: tables/rec_v2.tex
\begin{table*}[t]
   \centering
    %   \caption{Comparison with state-of-the-art methods in terms of Acc@0.5 on three benchmark datasets for the REC task. Following~\cite{yang2022lavt}, we refer to the language model of each reference method as the main learnable function that transforms word embeddings before multi-modal feature fusion. RN101 and RN152 denotes ResNet-101 and ResNet-152~\cite{he2016deep}, WRN101 refers to Wide ResNet-101~\cite{zagoruyko2016wide}, ENB3 denotes EfficientNetB3~\cite{enb3} and DN53 refers to Darknet-53~\cite{redmon2018yolov3}.}
   \scalebox{0.95}{\setlength{\tabcolsep}{2.0mm}{\begin{tabular}{l|l|l|c|c|c|c|c|c|c|c|c}
      \toprule[1pt]
      \multirow{2}{*}{Method} & Visual &
      Text & 
      \multicolumn{3}{c|}{RefCOCO}  & \multicolumn{3}{c|}{RefCOCO+} & \multicolumn{2}{c|}{RefCOCOg} & ReferIt\\
      \cline{4-12}
                                    & Backbone & Encoder & val   & test A & test B & val & test A & test B & val & test & test \\
      \hline
    %   MAttNet~\cite{yu2018mattnet} & RN101 & Bi-LSTM & 76.65 & 81.14 & 69.99 & 65.33 & 71.62 & 56.02 & 66.58 & 67.27 \\ 
      UNTIER-L~\cite{chen2020uniter} & RN101 & BERT & 81.41 & 87.04 & 74.17 & 75.90 & 81.45 & 66.70 & 74.86 & 75.77 & - \\
      VILLA-L~\cite{gan2020large} & RN101 & BERT & 82.39 & 87.48 & 74.84 & 76.17 & 81.54 & 66.84 & 76.18 & 76.71 & -\\
      RefTr~\cite{li2021referring} & RN101 & BERT-base & 85.65 & 88.73 & 81.16 & 77.55 & 82.26 & 68.99 & 79.25 & 80.01 & 76.18\\ 
    SeqTR~\cite{zhu2022seqtr} & DN53 & Bi-GRU & 87.00 & 90.15 & 83.59 & 78.69 & 84.51 & 71.87 & 82.69 & 83.37 & 69.66 \\
      MDETR~\cite{kamath2021mdetr} & ENB3 & RoBERTa-base & 87.51 & 90.40 & 82.67 & 81.13 & 85.52 & 72.96 & 83.35 & 83.31 & - \\
      OFA-B~\cite{wang2022unifying} & RN101 & 
      Embedding layer &  88.48 & 90.67 & 83.30 & 81.39 & 87.15 & 74.29 & 82.29 & 82.31 & - \\
    UniTAB~\cite{yang2022unitab} & RN101 & RoBERT-base & 88.59 & 91.06 & 83.75 & 80.97 & 85.36 & 71.55 & 84.58 & 84.70 & - \\
      OFA-L~\cite{wang2022unifying} & RN152 & Embedding layer & 90.05 & \textbf{92.93} & 85.26 & \textbf{85.80} & \textbf{89.87} & \textbf{79.22} & \textbf{85.89} & \textbf{86.55} & -  \\
      \hline
      %PolyFormer-B-1D & Swin-B & BERT-base & 89.53 & 91.92 & 86.52 & 84.26 & 88.11 & 77.71 & 85.01 & 85.60 \\
      %PolyFormer-B-2D & Swin-B & BERT-base & 89.63 & 91.62 & 86.63 & 83.84 & 88.51 & 77.01 & 84.48 & 84.96 \\
      %PolyFormer & Swin-B & BERT-base & 89.27 & 91.89 & 86.52 & 83.57 & 88.00 & 76.74 & 84.46 & 84.61 \\
      PolyFormer-B & Swin-B & BERT-base & 89.73 & 91.73 & 86.03 & 83.73 & 88.60 & 76.38 & 84.46 & 84.96 & 80.90 \\
      %\textbf{PloyFormer} & Swin-L & BERT-base & \textbf{90.35} & 92.81 & \textbf{87.20} & 84.6 & 89.07 & 78.30 & 85.78 & 86.10 \\
      \textbf{PolyFormer-L} & Swin-L & BERT-base & \textbf{90.38} & 92.89 & \textbf{87.16} & 84.98 & 89.77 & 77.97 & 85.83 & 85.91 & \textbf{81.50} \\
      %PolyFormer-L-1D & Swin-L & BERT-base & 89.87 & 92.15 & 86.77 & 84.28 & 88.28 & 77.75 & 85.13 & 85.24 \\
      %PolyFormer & Swin-L & BERT-base & 90.35 & 92.81 & 87.20 & 84.6 & 89.07 & 78.30 & 85.78 & 86.10 \\
      \bottomrule[1pt]
   \end{tabular}}}
   \caption{Comparison with the state-of-the-art methods on three referring expression comprehension benchmarks. ENB3 denotes EfficientNet-B3~\cite{tan2019efficientnet}.
%   OFA utilizes
% 19M image-text data for pre-training, while PolyFormer only uses 165K images. 
% \zc{if you want to highlight the best results, all best results should be highlighted if they are not polyformer.} \jl{updated}
}
\vspace{-1mm}
   \label{tab:rec}
\end{table*}

%% file: tables/rvs.tex
\begin{table}[t]
\centering
\scalebox{0.96}{\setlength{\tabcolsep}{1mm}{\begin{tabular}{l|c|c|c|c}
\toprule
Method            & Visual Backbone& $\mathcal{J}\&\mathcal{F} $ & $\mathcal{J} $   & $\mathcal{F}$    \\
\midrule
% CMSA~\cite{ye2019cross} & ResNet-50 & 34.7 & 32.2 & 37.2  \\
CMSA+RNN ~\cite{ye2019cross} & ResNet-50 & 40.2 & 36.9 & 43.5 \\
URVOS~\cite{seo2020urvos}              & ResNet-50 & 51.5  & 47.3  & 56.0     \\
CITD~\cite{liang2021rethinking}              & ResNet-101 & 56.4  & 54.8  & 58.1     \\
% CMPC-V~\cite{liu2021cross} & - & - & -  \\
% ClawCraneNet~\cite{liang2021clawcranenet} & - & - & -   \\
% MTTR ~\cite{botach2022mttr} & - & - & -  \\
ReferFormer~\cite{wu2022referformer}       & Swin-L & 60.5  & 57.6  & 63.4   \\
ReferFormer~\cite{wu2022referformer}       & Video-Swin-B & 61.1  & \textbf{58.1}  & 64.1   \\
\midrule
% PolyFormer$^*$ (Ours) & 60.8 & 56.3 & 65.3 &  69.9 \\
% PolyFormer-B-2D$^*$ (Ours)  & 58.8 & 54.6 & 63.0  \\
% \textbf{PolyFormer}$^*$ & \textbf{61.3 } & 57.1 & \textbf{65.5} \\
PolyFormer-B\dag & Swin-B & 60.9 & 56.6 & 65.2 \\
\textbf{PolyFormer-L}\dag & Swin-L &\textbf{61.5} & 57.2 & \textbf{65.8} \\
% PolyFormer-L-2D$^*$ (Ours)  & 61.3  & 57.1 & 65.5 \\
\bottomrule
\end{tabular}}}
\caption{Comparison with the state-of-the-art methods on Ref-DAVIS17. \dag means our model is trained on image datasets only. ReferFormer is trained on both image and video datasets. 
% \zc{put all tables/figures on the top of each page.}
}
\vspace{-2mm}
\label{tab:rvs-davis}
\end{table}

%% file: tables/regression.tex
\begin{table}[t]
\centering
\scalebox{0.92}{\setlength{\tabcolsep}{1.5mm}\setlength{\extrarowheight}{3pt}
{\begin{tabular}{l|l|c|c|c}
\toprule
\multicolumn{2}{c|}{Decoder}            &RefCOCO & RefCOCO+   & RefCOCOg    \\ 
\hline
% CMSA~\cite{ye2019cross} & 34.7 & 32.2 & 37.2  \\
% CMSA+RNN ~\cite{ye2019cross} & 40.2 & 36.9 & 43.5 \\
% URVOS~\cite{seo2020urvos}              & 51.5  & 47.3  & 56.0     \\
% CMPC-V~\cite{liu2021cross} & - & - & -  \\
% ClawCraneNet~\cite{liang2021clawcranenet} & - & - & -   \\
% MTTR ~\cite{botach2022mttr} & - & - & -  \\
\multirow{2}{*}{\rotatebox{90}{RIS}}& Classification       & 74.11  & 68.79  & 67.69   \\
\cline{2-5}
% PolyFormer$^*$ (Ours) & 60.8 & 56.3 & 65.3 &  69.9 \\
% PolyFormer-B-2D$^*$ (Ours)  & 58.8 & 54.6 & 63.0  \\
& \textbf{Regression} & \textbf{75.96}\small{ (+1.85)} & \textbf{70.65}\small{ (+1.86)}& \textbf{69.36}\small{ (+1.67)} \\ 
% PolyFormer-L-2D$^*$ (Ours)  & 61.3  & 57.1 & 65.5 \\
\hline
\multirow{2}{*}{\rotatebox{90}{REC}} & Classification       & 87.03 & 81.35  & 82.21   \\
\cline{2-5}
% PolyFormer$^*$ (Ours) & 60.8 & 56.3 & 65.3 &  69.9 \\
% PolyFormer-B-2D$^*$ (Ours)  & 58.8 & 54.6 & 63.0  \\
& \textbf{Regression} & \textbf{89.73}\small{ (+2.70)} & \textbf{83.73}\small{ (+2.38)}& \textbf{84.46}\small{ (+2.25)} \\
% PolyFormer-L-2D$^*$ (Ours)  & 61.3  & 57.1 & 65.5 \\
\bottomrule
\end{tabular}}}
\caption{
% Ablation study on classification \textit{v.s.} regression-based decoder.
Ablation study on regression-based decoder.
% \zc{I would say classification v.s. regression. Using PolyFormer, the readers don't see what you ablate here.}\hd{fixed}
} 
%\vspace{-2mm}
\label{tab:decoder}
\end{table}

%% file: figs/attn_v3.tex
%& $\textrm{XraySyn}_{ref}$ 
\captionsetup[subfigure]{labelformat=empty}
\begin{figure*}[t]
    \setlength{\tabcolsep}{2.5pt}
    \centering
    \small
    \begin{tabular}[b]{ccccccc}
        % \hline
       
        \multicolumn{7}{l}{Expression: ``an Asian girl with a pink shirt eating at the table"} \\
        \begin{subfigure}[b]{0.125\linewidth}
        \includegraphics[width=\textwidth]{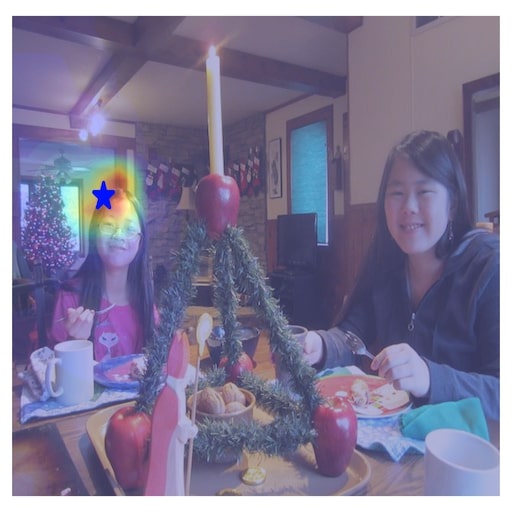}
        \caption{$t_{poly}=1$ (Start)}
        \end{subfigure} &
        \begin{subfigure}[b]{0.125\linewidth}
            \includegraphics[width=\textwidth]{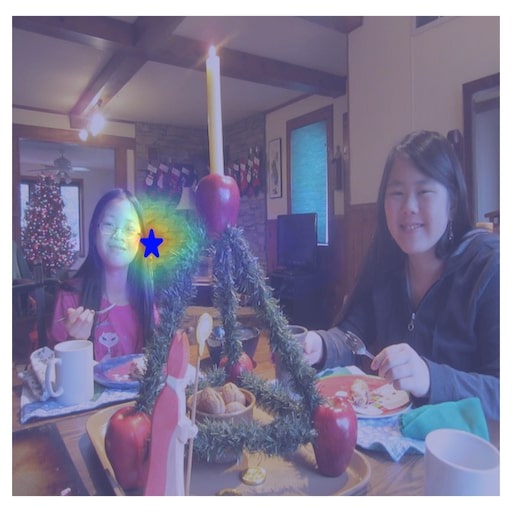}
        \caption{$t_{poly}=5$}
        \end{subfigure} &  
        \begin{subfigure}[b]{0.125\linewidth}
        \includegraphics[width=\textwidth]{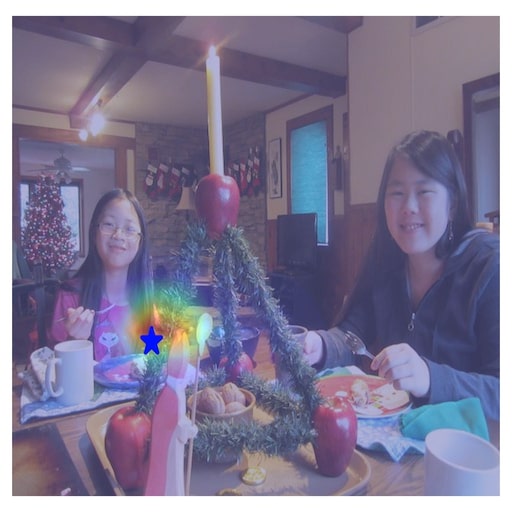}
        \caption{$t_{poly}=9$}
        \end{subfigure} &
        \begin{subfigure}[b]{0.125\linewidth}
        \includegraphics[width=\textwidth]{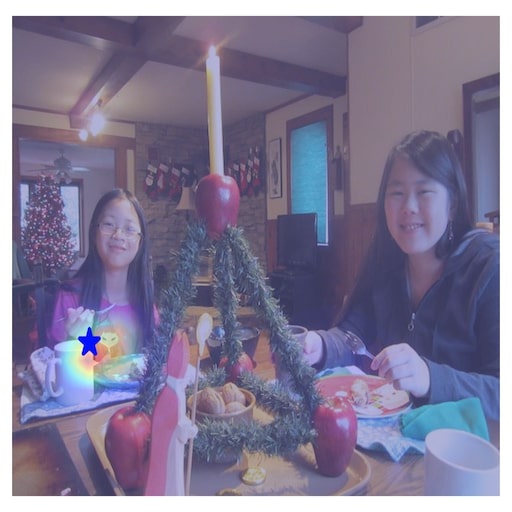}
        \caption{$t_{poly}=13$}
        \end{subfigure} &
        \begin{subfigure}[b]{0.125\linewidth}
        \includegraphics[width=\textwidth]{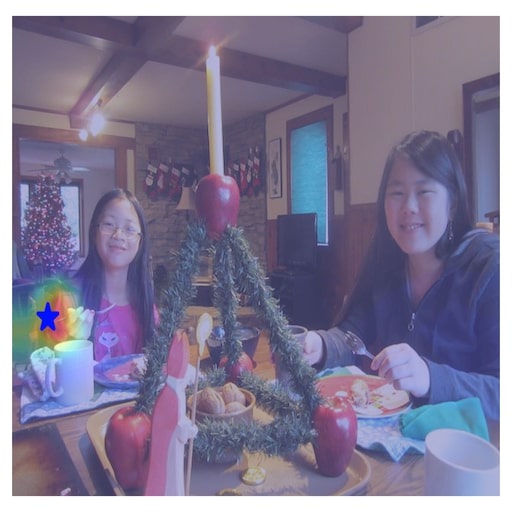}
        \caption{$t_{poly}=17$}
        \end{subfigure} &
        \begin{subfigure}[b]{0.125\linewidth}
        \includegraphics[width=\textwidth]{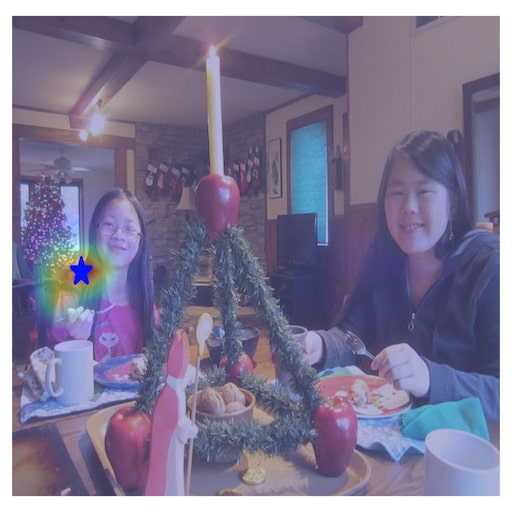}
        \caption{$t_{poly}=20$}
        \end{subfigure} &
        \begin{subfigure}[b]{0.125\linewidth}
        \includegraphics[width=\textwidth]{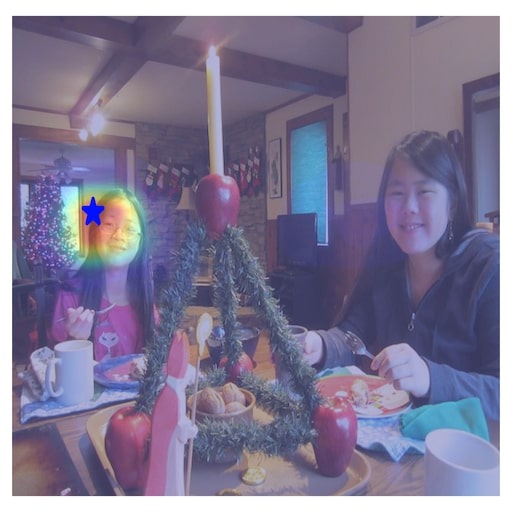}
        \caption{$t_{poly}=23$ (End)}
        \end{subfigure} \\

        % \multicolumn{6}{l}{Expression: ``a chili dog with slices of cheese visible under the chili"} \\
        % \begin{subfigure}[b]{0.13\linewidth}
        % \includegraphics[width=\textwidth]{figs/attn_v3/15_0_step_2.png}
        % \caption{$t_{poly}=0$}
        % \end{subfigure} &
        % \begin{subfigure}[b]{0.13\linewidth}
        % \includegraphics[width=\textwidth]{figs/attn_v3/15_0_step_13.png}
        % \caption{$t_{poly}=11$}
        % \end{subfigure} &
        % \begin{subfigure}[b]{0.13\linewidth}
        % \includegraphics[width=\textwidth]{figs/attn_v3/15_0_step_20.png}
        % \caption{$t_{poly}=18$}
        % \end{subfigure} &
        % \begin{subfigure}[b]{0.13\linewidth}
        % \includegraphics[width=\textwidth]{figs/attn_v3/15_0_step_29.png}
        % \caption{$t_{poly}=27$}
        % \end{subfigure} &
        % \begin{subfigure}[b]{0.13\linewidth}
        % \includegraphics[width=\textwidth]{figs/attn_v3/15_0_step_36.png}
        % \caption{$t_{poly}=34$}
        % \end{subfigure} &  
        % \begin{subfigure}[b]{0.13\linewidth}
        % \includegraphics[width=\textwidth]{figs/attn_v3/15_0_step_40.png}
        % \caption{$t_{poly}=38$}
        % \end{subfigure} \\ 

        % \hline
    \end{tabular}
    % \caption{Visualization of averaged cross-attention map in PolyFormer decoder at inference step $t_{poly}$ for polygon prediction. The last column corresponds to the last inference step.}
    %\caption{Decoder's cross-attention map when predicting the polygons. The first and last columns visualize the first and last predicted vertex coordinate tokens, respectively. $\star$ indicates the vertex prediction at time step $t_{poly}$.}
    \caption{The cross-attention maps of the decoder when generating the polygon. $\star$ is the 2D vertex prediction at inference step $t_{poly}$.}
    \label{fig:attn}
\end{figure*}

%% file: figs/vis.tex
%& $\textrm{XraySyn}_{ref}$ 
%\captionsetup[subfigure]{labelformat=empty}
\begin{figure*}[t]
    \centering
    \small
    \setlength{\tabcolsep}{0.8mm}
    {\begin{tabular}[b]{lccccccc}
        % \hline
        \rotatebox{90}{\hskip 2em LAVT} &
        
        \begin{subfigure}[b]{0.1239\linewidth}
        \includegraphics[width=\textwidth]{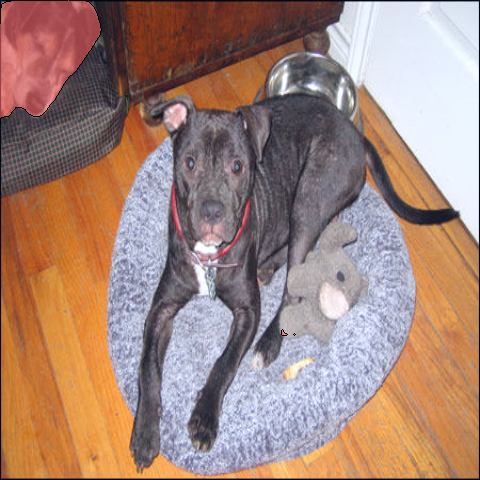}
        \end{subfigure} &  
        
        \begin{subfigure}[b]{0.1239\linewidth}
            \includegraphics[width=\textwidth]{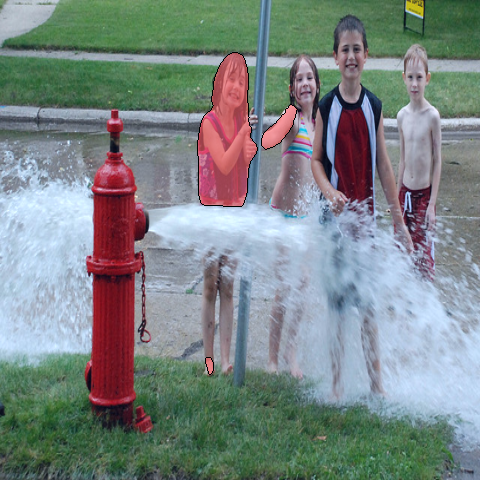}
        \end{subfigure} &
        \begin{subfigure}[b]{0.1239\linewidth}
            \includegraphics[width=\textwidth]{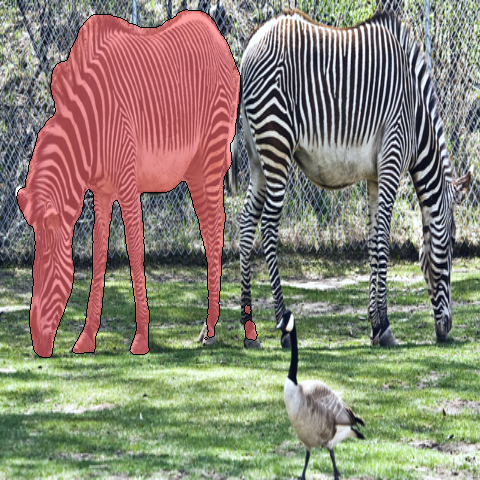}
        \end{subfigure} &
        \begin{subfigure}[b]{0.1239\linewidth}
            \includegraphics[width=\textwidth]{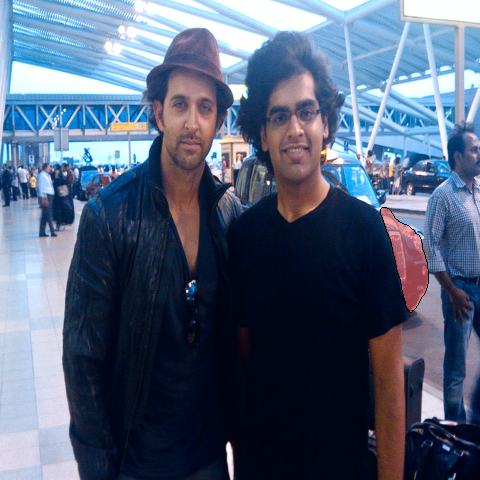}
        \end{subfigure} &
        \begin{subfigure}[b]{0.1239\linewidth}
        \includegraphics[width=\textwidth]{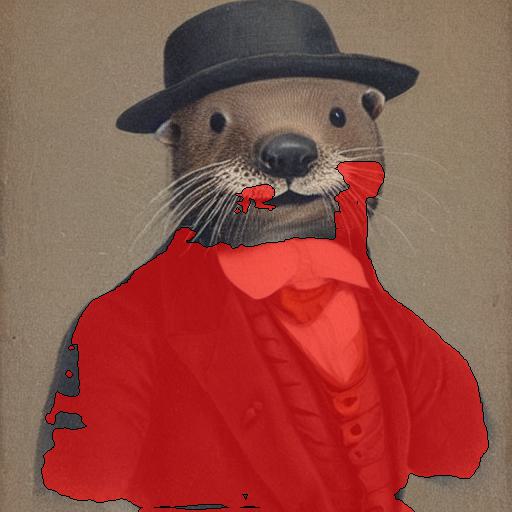}
        \end{subfigure} &
        \begin{subfigure}[b]{0.1239\linewidth}
            \includegraphics[width=\textwidth]{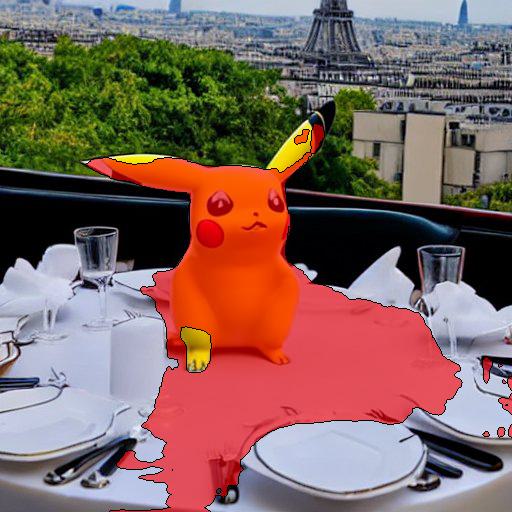}
        \end{subfigure} &  
        \begin{subfigure}[b]{0.1239\linewidth}
            \includegraphics[width=\textwidth]{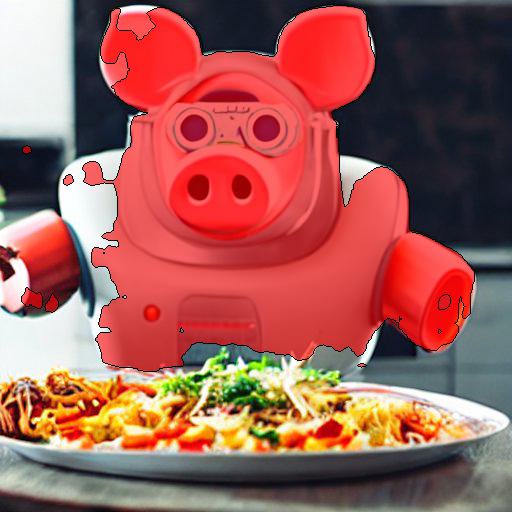}
        \end{subfigure}\\ 
        
        \rotatebox{90}{\hskip 2em SeqTR} &
                \begin{subfigure}[b]{0.1239\linewidth}
            \includegraphics[width=\textwidth]{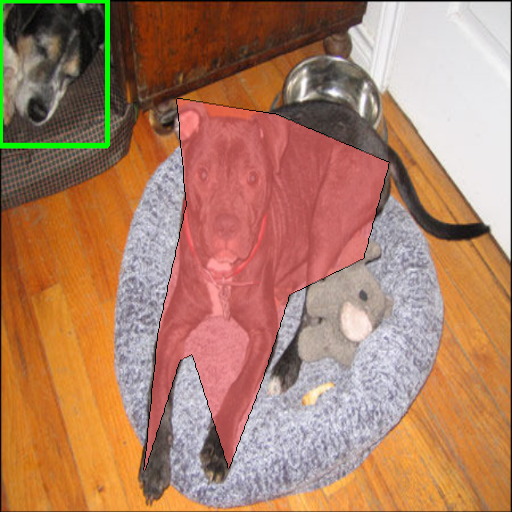}
        \end{subfigure} &  
        \begin{subfigure}[b]{0.1239\linewidth}
            \includegraphics[width=\textwidth]{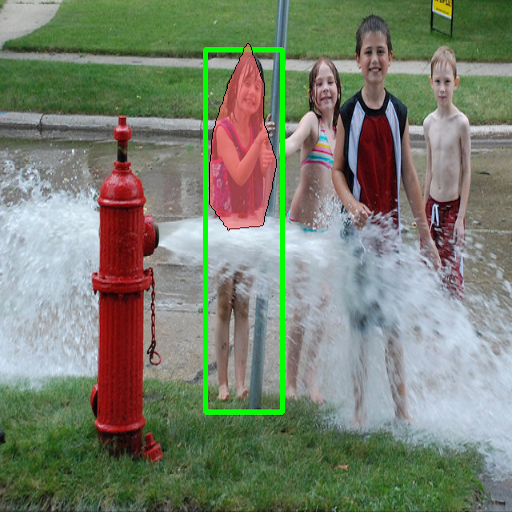}
        \end{subfigure} &
        \begin{subfigure}[b]{0.1239\linewidth}
            \includegraphics[width=\textwidth]{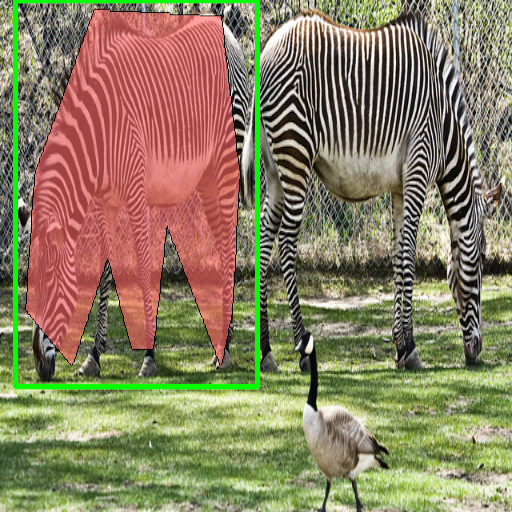}
        \end{subfigure} &
        \begin{subfigure}[b]{0.1239\linewidth}
            \includegraphics[width=\textwidth]{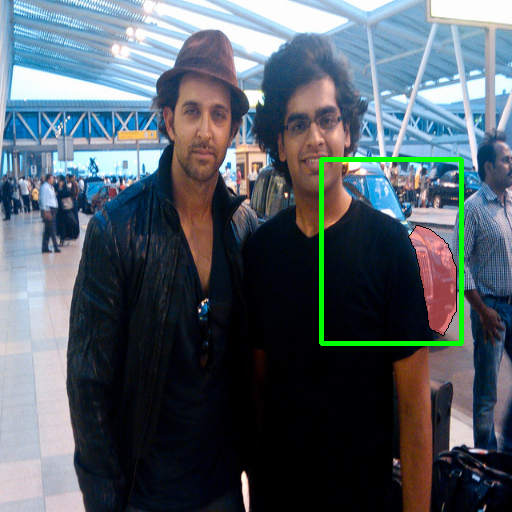}
        \end{subfigure} &
        \begin{subfigure}[b]{0.1239\linewidth}
        \includegraphics[width=\textwidth]{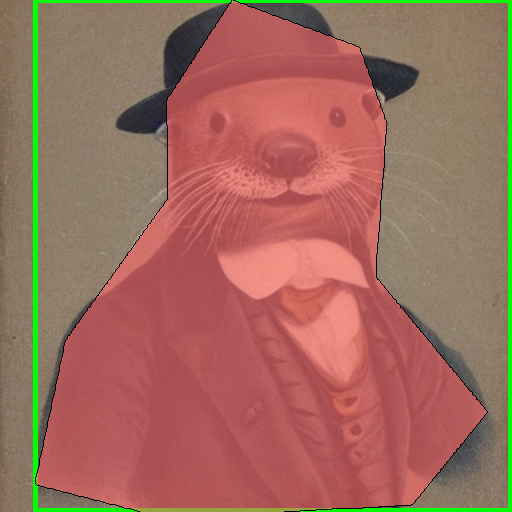}
        \end{subfigure} &
        \begin{subfigure}[b]{0.1239\linewidth}
            \includegraphics[width=\textwidth]{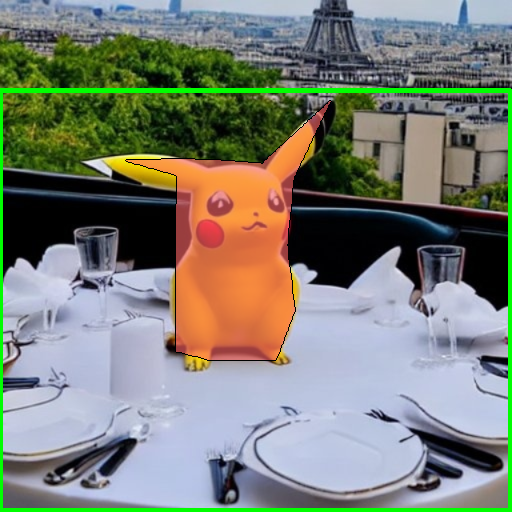}
        \end{subfigure} &  
        % \begin{subfigure}[b]{0.12\linewidth}
        %     \includegraphics[width=\textwidth]{figs/zero_shot/cabin_seqtr.png}
        % \end{subfigure} &
        % \begin{subfigure}[b]{0.12\linewidth}
        %     \includegraphics[width=\textwidth]{figs/zero_shot/shiba_seqtr.png}
        % \end{subfigure} &
        \begin{subfigure}[b]{0.1239\linewidth}
            \includegraphics[width=\textwidth]{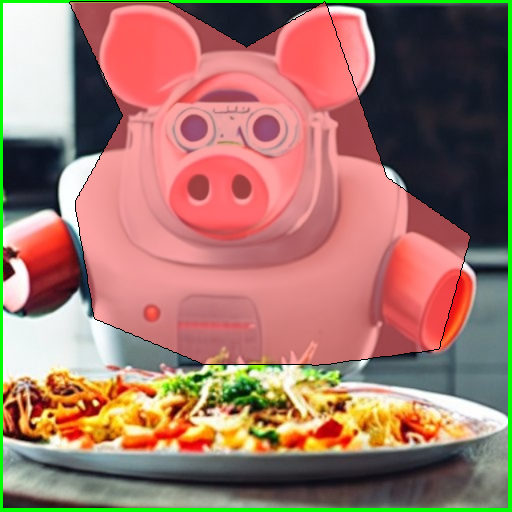}
        \end{subfigure}\\

        \rotatebox{90}{\hskip 1.5em  \textbf{PolyFormer}}&
        \begin{subfigure}[t]{0.1239\linewidth}
            \includegraphics[width=\textwidth]{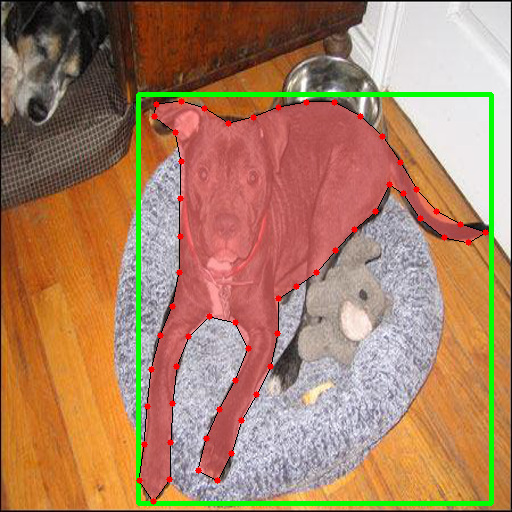}
            \caption{(a) ``a dark grey dog on a light grey round bed wearing a red collar"}
        \end{subfigure} &  
        \begin{subfigure}[t]{0.1239\linewidth}
            \includegraphics[width=\textwidth]{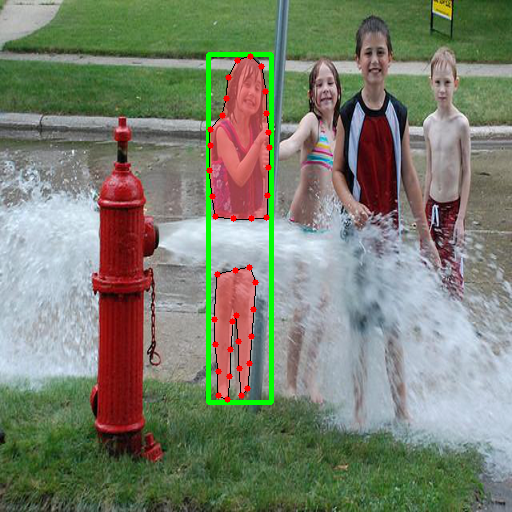}
            \caption{(b) ``girl in purple"}
        \end{subfigure} &
        \begin{subfigure}[t]{0.1239\linewidth}
            \includegraphics[width=\textwidth]{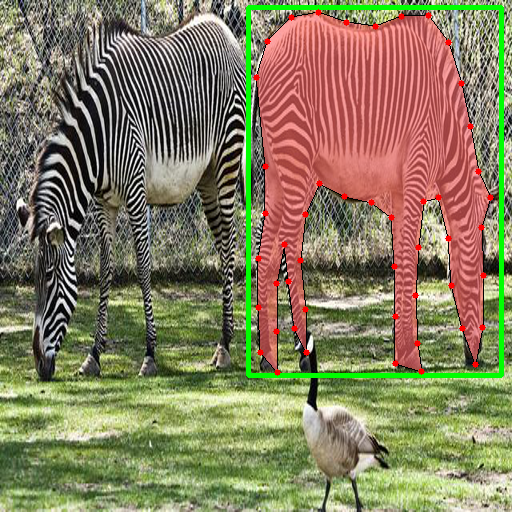}
            \caption{(c) ``zebra eating grass with a goose in front of it"}
        \end{subfigure} &
        \begin{subfigure}[t]{0.1239\linewidth}
            \includegraphics[width=\textwidth]{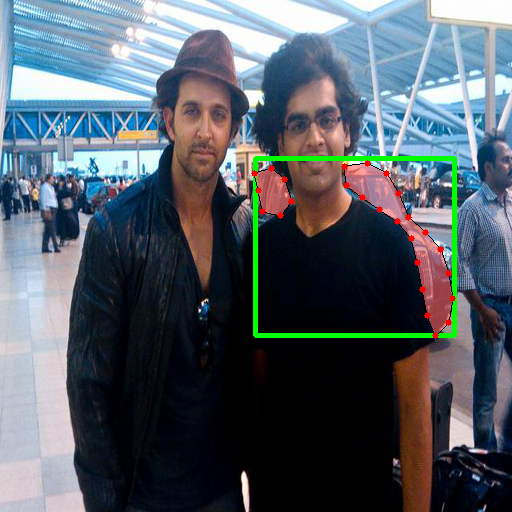}
            \caption{(d) ``a black car parked at a transportation terminal"}
        \end{subfigure} &

        \begin{subfigure}[t]{0.1239\linewidth}
        \includegraphics[width=\textwidth]{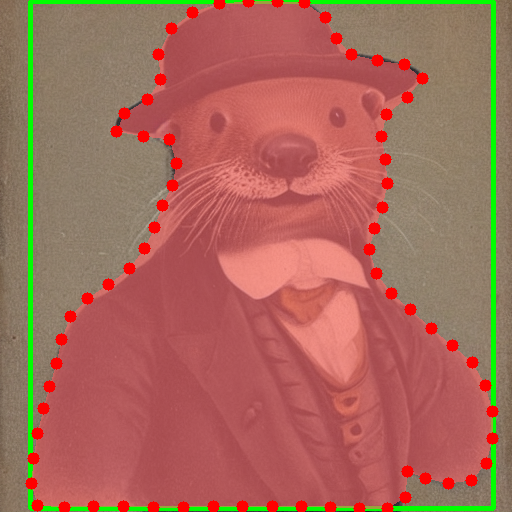}
        \caption{(e) ``A gentleman otter in a 19th century portrait"}
        \end{subfigure} &
        \begin{subfigure}[t]{0.1239\linewidth}
            \includegraphics[width=\textwidth]{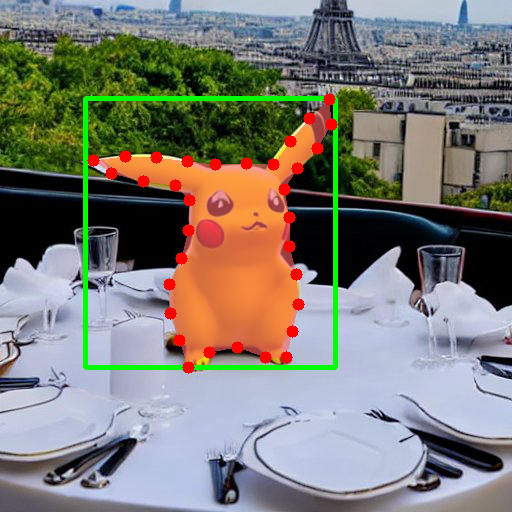}
            \caption{(f) ``A pikachu fine-dining with a view to the Eiffel Tower"}
        \end{subfigure} &  
        % \begin{subfigure}[t]{0.12\linewidth}
        %     \includegraphics[width=\textwidth]{figs/zero_shot/cabin_poly_2d_large.png}
        %     \caption{``A small cabin on top of a snowy mountain in the style of Disney artstation"}
        % \end{subfigure} &
        % \begin{subfigure}[t]{0.12\linewidth}
        %     \includegraphics[width=\textwidth]{figs/zero_shot/shiba_poly_2d_large.png}
        %     \caption{``A shiba inu puppy painted by Monet"}
        % \end{subfigure} &
        \begin{subfigure}[t]{0.1239\linewidth}
            \includegraphics[width=\textwidth]{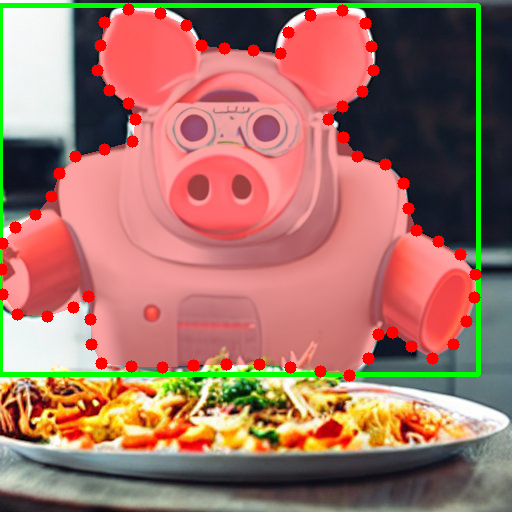}
            \caption{(g) ``A pig robot preparing a delicious meal"}
        \end{subfigure}\\ 
        %Input Image & LAVT~\cite{yang2022lavt} & SeqTR~\cite{zhu2022seqtr} & PolyFormer (ours)  &  Ground-truth \\

        % \hline
    \end{tabular}}
    \caption{
    % Visual comparisons between LAVT~\cite{yang2022lavt}, SeqTR~\cite{zhu2022seqtr} and PolyFormer. (a)-(d): RefCOCOg test set; (e)-(g): synthetic images generated by Stable Diffusion~\cite{rombach2021highresolution}.
    The results of LAVT~\cite{yang2022lavt} (top),
    SeqTR~\cite{zhu2022seqtr} (middle), and PolyFormer (bottom) on RefCOCOg test set (a-d) and images generated by~\cite{rombach2021highresolution} (e-g). LAVT is for referring image segmentation only. For SeqTR, we generate the bounding boxes and segmentation masks from the task-specific models as they perform better than the multi-task variant.
    \iffalse
    PolyFormer simultaneously predicts the bounding box and polygon vertices that forms the segmentation mask. 
    \fi
    % Blue point indicates the starting point of the vertex sequence. 
    \iffalse
    LAVT is for referring image segmentation only. For SeqTR, we generate the bounding boxes and segmentation masks from the task-specific models as they perform better than the multi-task model.
    \fi
    }
    % \caption{Visualization results on RefCOCOg. Our unified framework is able to detect and segment the referred object simultaneously. }
    \label{fig:visualization}
    %\vspace{-1em}
\end{figure*}

% \begin{tblr}{
%   colspec = {Q[b,4em]|Q[c,m]}
% }
%   LAVT~\cite{yang2022lavt} & \includegraphics[width=0.18\textwidth]{figs/visualization_2/449_1_lavt.png}\\
% \end{tblr}

%% file: appendix.tex
\paragraph{Limitations and Broader Impacts.}
% \subsection{Limitations and Broader Impacts}
The training of PolyFormer requires accurate bounding box and polygon annotations. How to reduce such dependence and utilize weakly-supervised data for region-level image understanding needs further exploration. For the data and model, we need to further understand the broader impacts including but not limited to fairness, social bias and potential misuse.

%%%%%%%%% BODY TEXT
% hyper-parameter

%     classification weight
    
%     det weight for multi-task

%  1d: 500, 2000 bins
 
%  2d: 32, 128 bins
 
%  1d classification on other datasets, refcoco test, refcoco+ test, refcocog test
% dataset details
 
%  seqtr data on combined dataset, results on refcoco val, refcoco+ val, refcocog val
 
%  failure case visualization and discussion: 
%  1) complete failure: find the wrong object; fail to predict any object.
%  2) partial failure: fail to find multiple parts; segment parts that does not belong to the target

% more attention visualizations

% classification vs regression decoder: base and large

% 1d vs 2d models: base and large

% \section*{Supplementary Material}
\section{Additional Dataset Details}
We evaluate PolyFormer on four benchmark image datasets, RefCOCO \cite{yu2016modeling}, RefCOCO+ \cite{yu2016modeling}, RefCOCOg \cite{mao2016generation, nagaraja2016modeling}, and ReferIt~\cite{referit}. All images of RefCOCO, RefCOCO+, and RefCOCOg are from the MS COCO dataset \cite{mscoco} and annotated with referring expressions. We further evaluate PolyFormer models for the Referring Video Object Segmentation (R-VOS) task on Ref-DAVIS17~\cite{khoreva2018video}.

\paragraph{\textbf{RefCOCO/RefCOCO+}:} These two datasets are collected using a two-player game\cite{yu2016modeling}. RefCOCO has 142,209 annotated expressions for 50,000 objects in 19,994 images, and RefCOCO+ consists of 141,564 expressions for 49,856 objects in 19,992 images. These two datasets are splitted into training, validation, test A and test B sets, where test A contains images of multiple people and test B contains images of multiple instances of all other objects. Compared to RefCOCO, location words are banned from the referring expressions in RefCOCO+, which makes it more challenging. 

\paragraph{\textbf{RefCOCOg}:} This dataset is collected on Amazon Mechanical Turk, where workers are asked to write natural language referring expressions for objects. RefCOCOg consists of 85,474 referring expressions for 54,822
objects in 26,711 images. RefCOCOg has longer, more complex expressions (8.4 words on average), while the expressions in RefCOCO and RefCOCO+ are more succinct (3.5 words on average), which makes RefCOCOg particularly challenging. We use the UMD partition~\cite{nagaraja2016modeling} for RefCOCOg as it provides both validation and testing sets and there is no overlapping between training and validation images. 

\paragraph{\textbf{ReferIt}:} ReferIt contains 130,364 referring expressions for 99,296 objects in 19,997 images collected from the SAIAPR-12 dataset~\cite{ESCALANTE2010419}. We use the cleaned
Berkeley split of the dataset, which consists of 58,838, 6,333, and 65,193 referring expressions in train, validation, and test sets, respectively. Compared to RefCOCO, RefCOCO+ and RefCOCOg, ReferIt contains more stuff segmentation masks, \textit{e.g.}, sky, ground. 

\paragraph{\textbf{Ref-DAVIS17}:} Ref-DAVIS17 contains 90 videos from the DAVIS17~\cite{pont20172017} dataset, where language descriptions are provided for specific objects in each video. It contains 1,544 referring expressions for 205 objects. The dataset is split into a training set and a validation set, containing 60 and 30 videos respectively. For each referred object, each of the two annotators  provides the descriptions of the first-frame and the full-video. For the Ref-DAVIS17 dataset, we use the standard evaluation metrics:  Region Jaccard ($\mathcal{J}$), Boundary F measure ($\mathcal{F}$), and their average value ($\mathcal{J}\&\mathcal{F}$).

\section{Additional Implementation Details}
The dimension of image feature
$C_v$ is 1024 for PolyFormer-B and 1536 for PolyFormer-L. The dimensions of language feature $C_l$ and coordinate embedding $C_e$ are 768. We use a linear layer to project the language and image features into the same dimension of 768. We adopt 12 attention heads in the self-attention and cross-attention layers, and GELU activations in the transformer encoder and decoder layers. For $L_{cls}$, we set the label smoothing factor to 0.1.

%\section{Additional Ablation Studies}
%In this section, we perform additional ablation studies to study the effects of several components of PolyFormer. 
% All models are based on Swin-B visual backbone with 6 encoder and decoder layers unless stated otherwise 
%All ablation experiments are performed using PolyFormer-B and we report the mIoU results for the referring image segmentation task unless stated otherwise.
% \zc{no experiments on Large model, right? if so, remove ``unless stated otherwise''}
% \section{Effect of number of bins}
\section{Additional Experiment Results}
To obtain the accurate coordinate embedding, we build a 2D coordinate codebook, $\mathcal{D}\in \mathbb{R}^{B_{H} \times B_{W} \times C_e}$, where $B_{H}$ and $B_{W}$ are the numbers of bins along the height and width dimensions, respectively. We train PolyFormer-B models with different number of bins $B_{H} \times B_{W}$ and the results are summarized in Table \ref{tab:ablation-bins}. We observe that using coordinate book with $64 \times 64$ bins achieves the best result, which is adopted by default in all the other experiments. %Reducing or increasing the number of bins both degrades model performance. 

\begin{table}[t]
\centering
\scalebox{1}{\setlength{\tabcolsep}{1.0mm}{\begin{tabular}{c|c|c|c}
\toprule
{$B_H\times B_W$}            & {RefCOCO} & {RefCOCO+} & {RefCOCOg} \\
\midrule
$32\times 32$  &75.07 & 70.15 & 68.49        \\ 
$64\times 64$  &\textbf{75.96} & \textbf{70.65} & \textbf{69.36}       \\ 
$128\times 128$  & 74.99 & 70.01 & 68.69        \\ 
\bottomrule
\end{tabular}}}
\caption{Ablation study on the size of 2D coordinate codebook.}
\label{tab:ablation-bins}
\end{table}

\section{More Visualization Results}
\subsection{Cross-attention Map} 
More cross-attention map visualization is shown in~\cref{fig:attn_supp}. We observe that the cross-attention map concentrates on the object referred by the sentence, and moves around the object boundary during the polygon generation process. 

\subsection{Prediction Visualization}
 \cref{fig:zero-shot_supp} shows more examples on the synthetic images generated by Stable Diffusion~\cite{rombach2021highresolution}. \cref{fig:visualization_supp} shows more examples on the RefCOCOg test set. It can be seen that PolyFormer is able to segment the referred object in challenging scenarios, \textit{e.g.}, instances with occlusion and complex shapes, instances that are partially displayed or require complex language understanding. In addition, PolyFormer demonstrates good generalization ability on synthetic images and text descriptions that have never been seen during training. In contrast, the state-of-the-arts LAVT \cite{yang2022lavt} and SeqTR \cite{zhu2022seqtr} fail to generate satisfactory results.

\input{figs/attn_supp}
\input{figs/zeroshot_supp}
\input{figs/visualization_supp}

%% file: figs/attn_supp.tex
%& $\textrm{XraySyn}_{ref}$ 
\captionsetup[subfigure]{labelformat=empty}
\begin{figure*}[t]
    \setlength{\tabcolsep}{1pt}
    \centering
    \small
    \begin{tabular}[b]{ccccccc}
        
        \multicolumn{7}{l}{Expression: ``a without hairy brown color teddy bear"} \\
        \begin{subfigure}[b]{0.125\linewidth}
        \includegraphics[width=\textwidth]{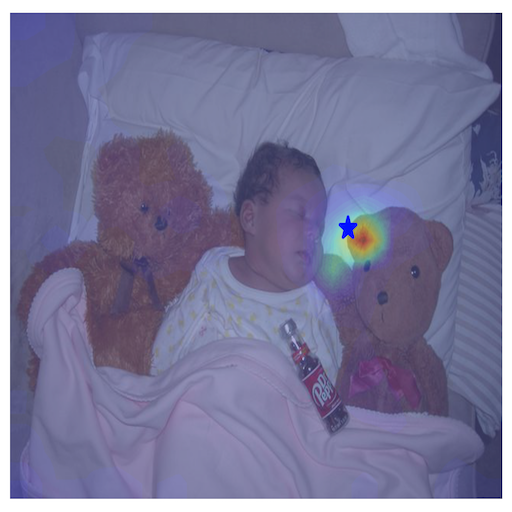}
        \caption{$t_{poly}=1$ (start)}
        \end{subfigure} &
        \begin{subfigure}[b]{0.125\linewidth}
        \includegraphics[width=\textwidth]{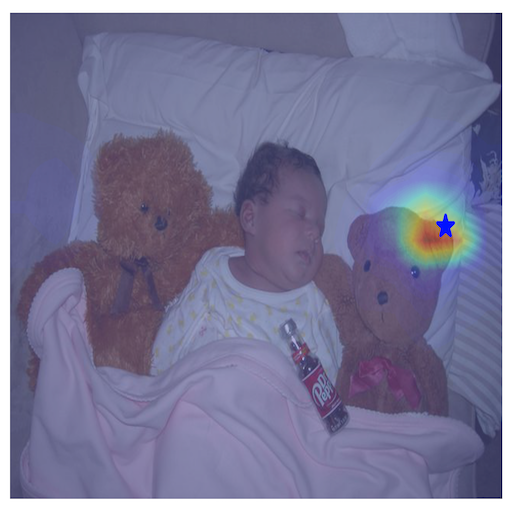}
        \caption{$t_{poly}=6$}
        \end{subfigure} &
        \begin{subfigure}[b]{0.125\linewidth}
        \includegraphics[width=\textwidth]{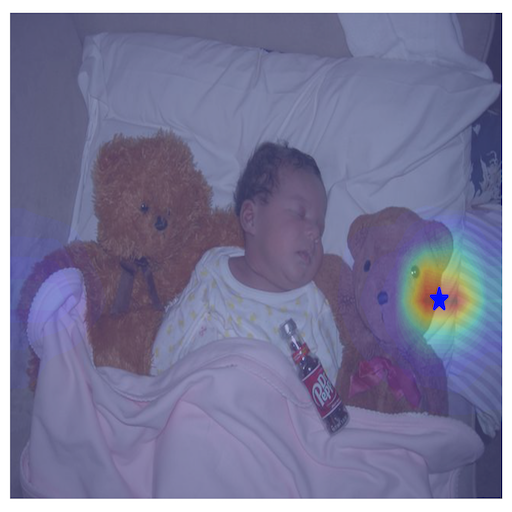}
        \caption{$t_{poly}=9$}
        \end{subfigure} &
        \begin{subfigure}[b]{0.125\linewidth}
        \includegraphics[width=\textwidth]{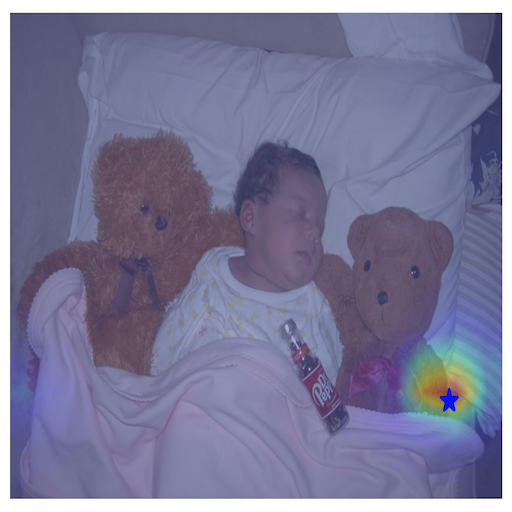}
        \caption{$t_{poly}=13$}
        \end{subfigure} &
        \begin{subfigure}[b]{0.125\linewidth}
        \includegraphics[width=\textwidth]{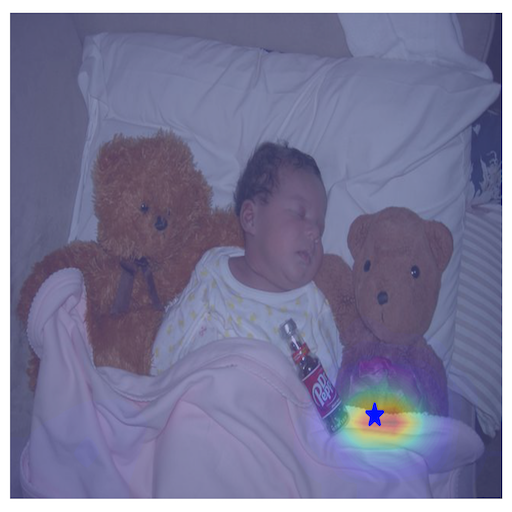}
        \caption{$t_{poly}=17$}
        \end{subfigure} &  
        \begin{subfigure}[b]{0.125\linewidth}
        \includegraphics[width=\textwidth]{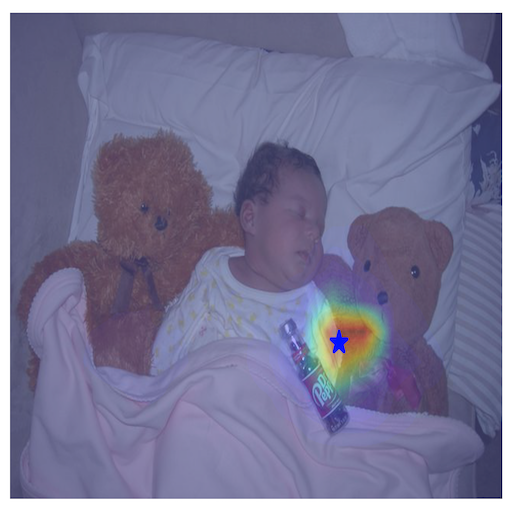}
        \caption{$t_{poly}=21$}
        \end{subfigure} &  
        \begin{subfigure}[b]{0.125\linewidth}
        \includegraphics[width=\textwidth]{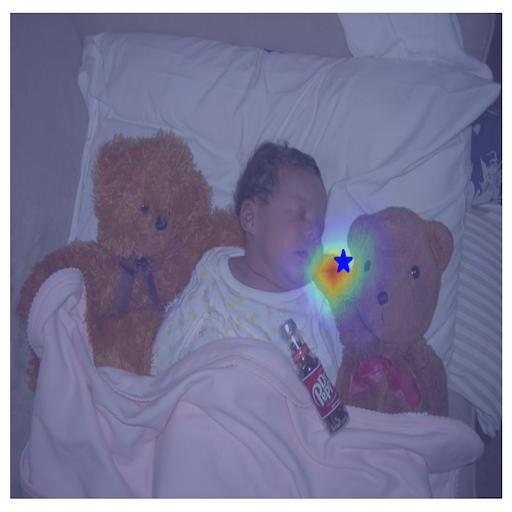}
        \caption{$t_{poly}=24$ (end)}
        \end{subfigure}\\ 
        
        \multicolumn{7}{l}{Expression: ``a chili dog with slices of cheese visible under the chili"} \\
        \begin{subfigure}[b]{0.125\linewidth}
        \includegraphics[width=\textwidth]{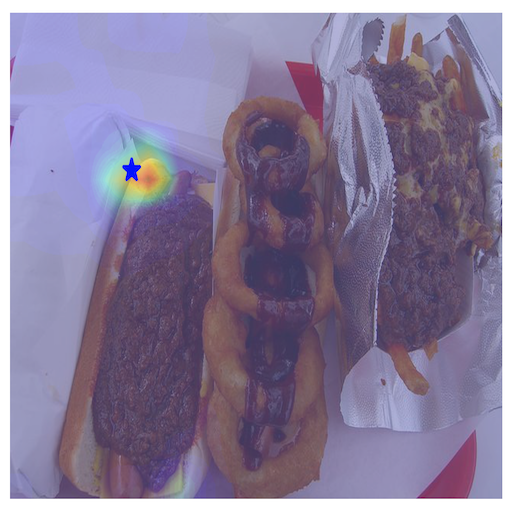}
        \caption{$t_{poly}=1$ (start)}
        \end{subfigure} &
        \begin{subfigure}[b]{0.125\linewidth}
        \includegraphics[width=\textwidth]{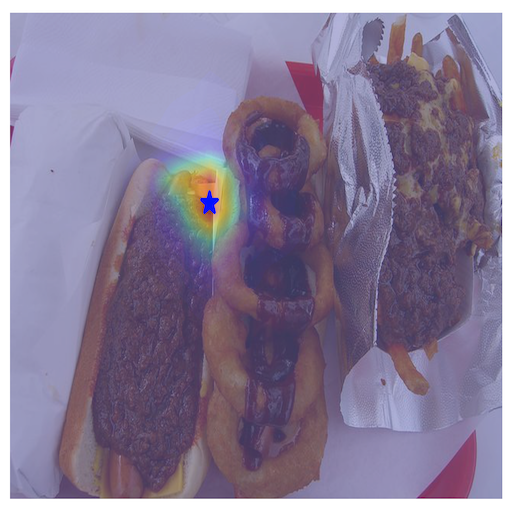}
        \caption{$t_{poly}=6$}
        \end{subfigure} &
        \begin{subfigure}[b]{0.125\linewidth}
        \includegraphics[width=\textwidth]{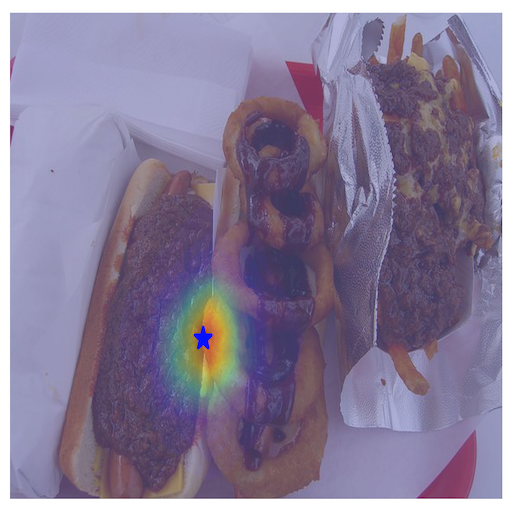}
        \caption{$t_{poly}=11$}
        \end{subfigure} &
        \begin{subfigure}[b]{0.125\linewidth}
        \includegraphics[width=\textwidth]{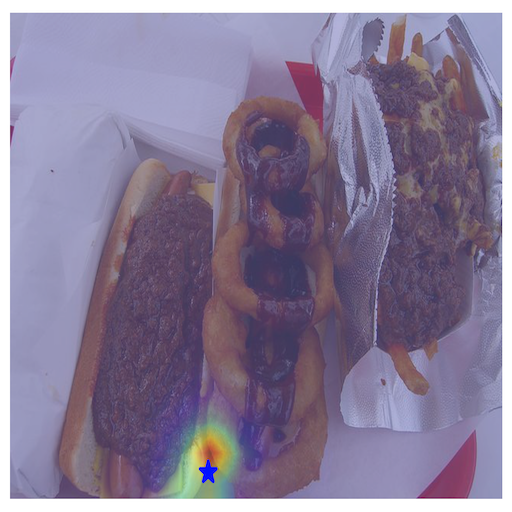}
        \caption{$t_{poly}=16$}
        \end{subfigure} &
        \begin{subfigure}[b]{0.125\linewidth}
        \includegraphics[width=\textwidth]{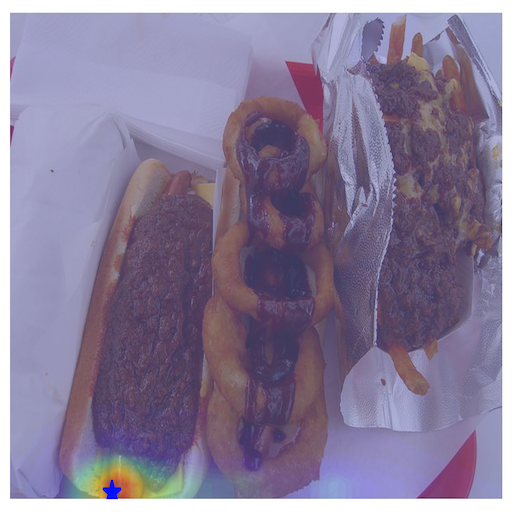}
        \caption{$t_{poly}=21$}
        \end{subfigure} &  
        \begin{subfigure}[b]{0.125\linewidth}
        \includegraphics[width=\textwidth]{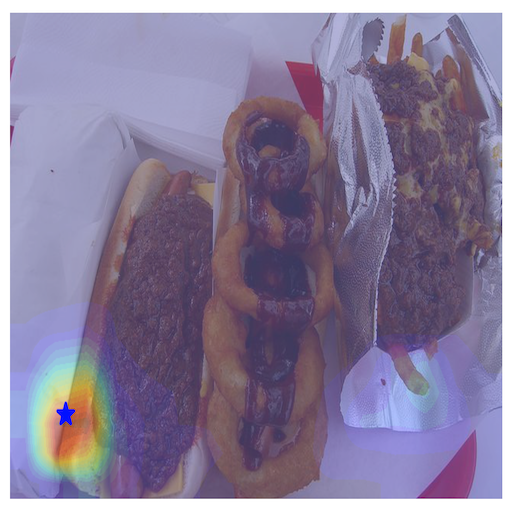}
        \caption{$t_{poly}=26$}
        \end{subfigure} &  
        \begin{subfigure}[b]{0.125\linewidth}
        \includegraphics[width=\textwidth]{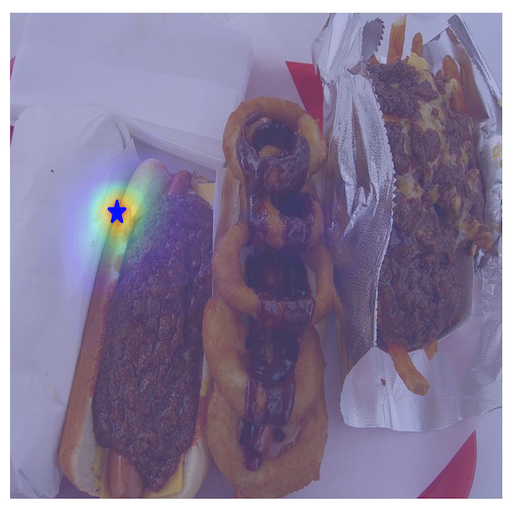}
        \caption{$t_{poly}=34$ (end)}
        \end{subfigure}\\ 
        
        \multicolumn{7}{l}{Expression: ``the orange closest to the banana"} \\
        \begin{subfigure}[b]{0.125\linewidth}
        \includegraphics[width=\textwidth]{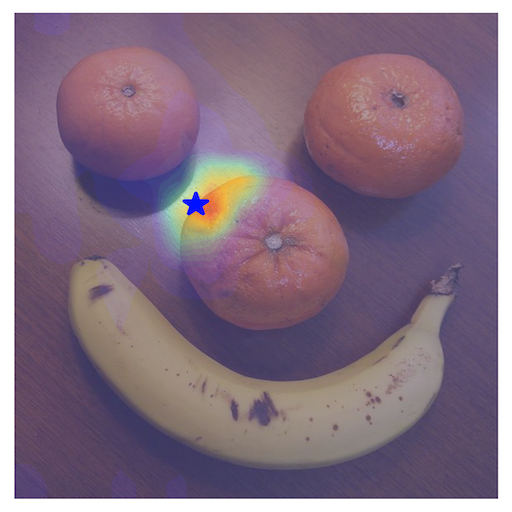}
        \caption{$t_{poly}=1$ (start)}
        \end{subfigure} &
        \begin{subfigure}[b]{0.125\linewidth}
        \includegraphics[width=\textwidth]{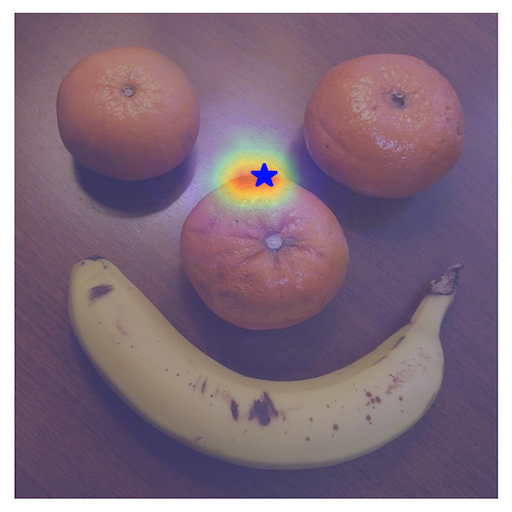}
        \caption{$t_{poly}=4$}
        \end{subfigure} &
        \begin{subfigure}[b]{0.125\linewidth}
        \includegraphics[width=\textwidth]{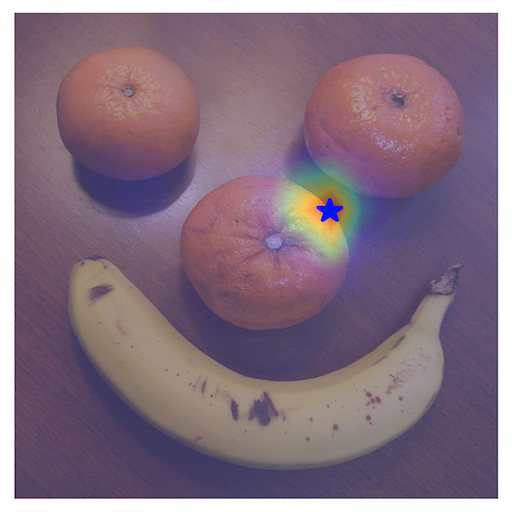}
        \caption{$t_{poly}=7$}
        \end{subfigure} &
        \begin{subfigure}[b]{0.125\linewidth}
        \includegraphics[width=\textwidth]{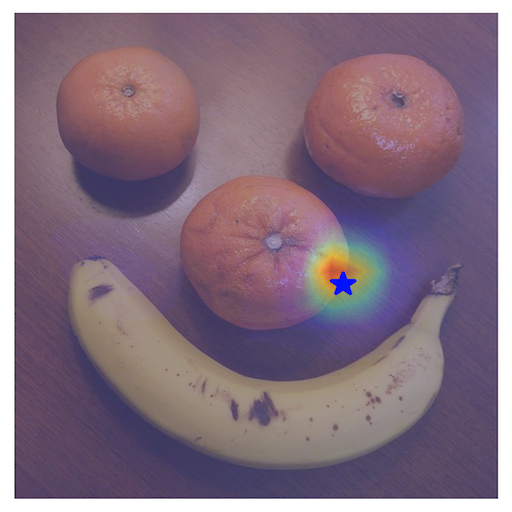}
        \caption{$t_{poly}=10$}
        \end{subfigure} &
        \begin{subfigure}[b]{0.125\linewidth}
        \includegraphics[width=\textwidth]{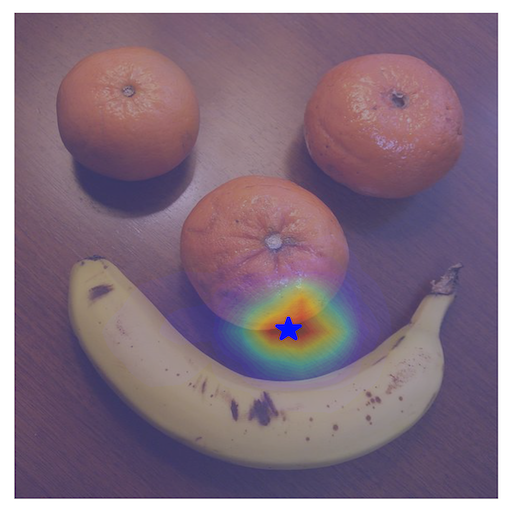}
        \caption{$t_{poly}=13$}
        \end{subfigure} &  
        \begin{subfigure}[b]{0.125\linewidth}
        \includegraphics[width=\textwidth]{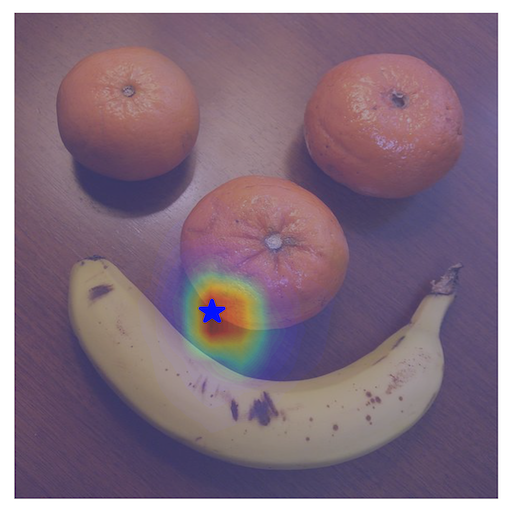}
        \caption{$t_{poly}=16$}
        \end{subfigure} &  
        \begin{subfigure}[b]{0.125\linewidth}
        \includegraphics[width=\textwidth]{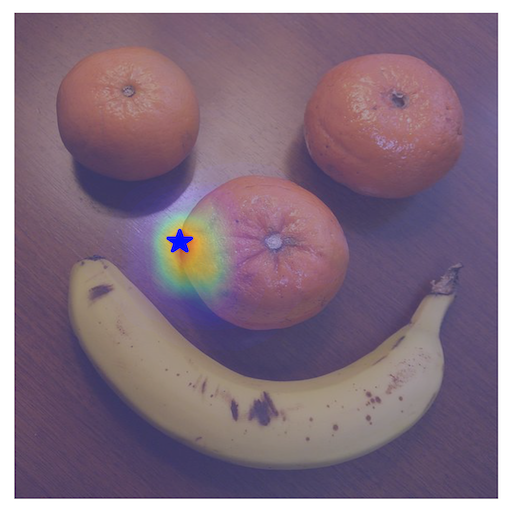}
        \caption{$t_{poly}=19$ (end)}
        \end{subfigure}\\ 

        % \hline
    \end{tabular}
    % \caption{Visualization of averaged cross-attention map in PolyFormer decoder at inference step $t_{poly}$ for polygon prediction. The last column corresponds to the last inference step.}
    \caption{Decoder's cross-attention map when predicting the polygons. $\star$ indicates the vertex prediction at time step $t_{poly}$.}
    \label{fig:attn_supp}
\end{figure*}

%% file: figs/zeroshot_supp.tex
%& $\textrm{XraySyn}_{ref}$ 
\captionsetup[subfigure]{labelformat=empty}
\begin{figure*}[t]
    \setlength{\tabcolsep}{0.5pt}
    \centering
    \small
    \setlength{\tabcolsep}{0.8mm}{\begin{tabular}[b]{lcccccc}
        % \hline
         
        %\multicolumn{5}{l}{Expression: ``a cow that is \textit{\textbf{being fed}} by a bottle"} \\
        
        % \rotatebox{90}{\hskip 3em Input}&
        % \begin{subfigure}[b]{0.15\linewidth}
        % \includegraphics[width=\textwidth]{figs/zero_shot/a gentleman otter in a 19th century portrait.png}
        % \end{subfigure} &
        % \begin{subfigure}[b]{0.15\linewidth}
        %     \includegraphics[width=\textwidth]{figs/zero_shot/A pikachu fine dining with a view to the Eiffel Tower.jpeg}
        % \end{subfigure} &  
        % \begin{subfigure}[b]{0.15\linewidth}
        %     \includegraphics[width=\textwidth]{figs/zero_shot/A small cabin on top of a snowy mountain in the style of Disney artstation.jpeg}
        % \end{subfigure} &
        % \begin{subfigure}[b]{0.15\linewidth}
        %     \includegraphics[width=\textwidth]{figs/zero_shot/A shiba inu puppy painted by monet.jpeg}
        % \end{subfigure} &
        % \begin{subfigure}[b]{0.15\linewidth}
        %     \includegraphics[width=\textwidth]{figs/zero_shot/pig.jpeg}
        % \end{subfigure}\\ 
        
        \rotatebox{90}{\hskip 3em LAVT} &
        \begin{subfigure}[b]{0.15\linewidth}
        \includegraphics[width=\textwidth]{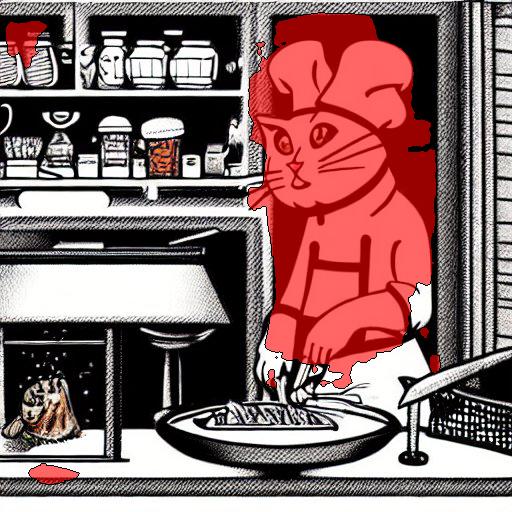}
        \end{subfigure} &
        \begin{subfigure}[b]{0.15\linewidth}
            \includegraphics[width=\textwidth]{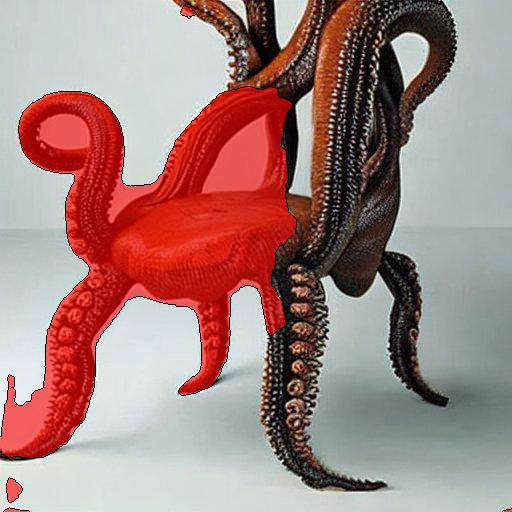}
        \end{subfigure} &  
        \begin{subfigure}[b]{0.15\linewidth}
            \includegraphics[width=\textwidth]{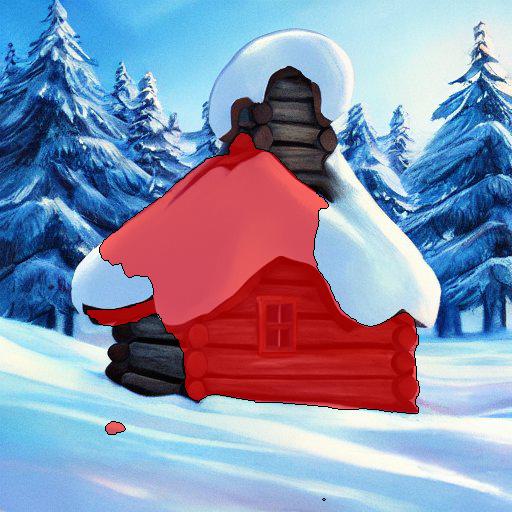}
        \end{subfigure} &
        \begin{subfigure}[b]{0.15\linewidth}
            \includegraphics[width=\textwidth]{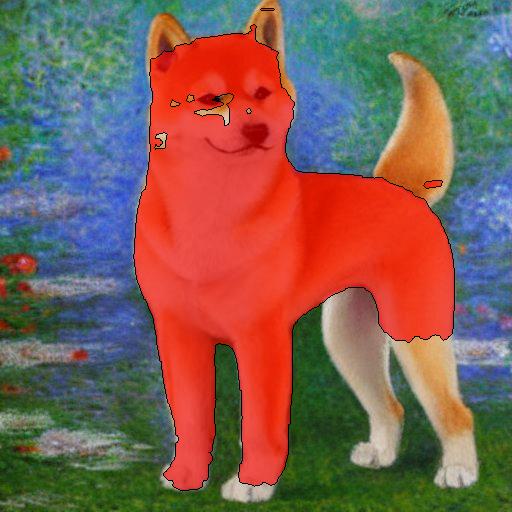}
        \end{subfigure} &
        \begin{subfigure}[b]{0.15\linewidth}
            \includegraphics[width=\textwidth]{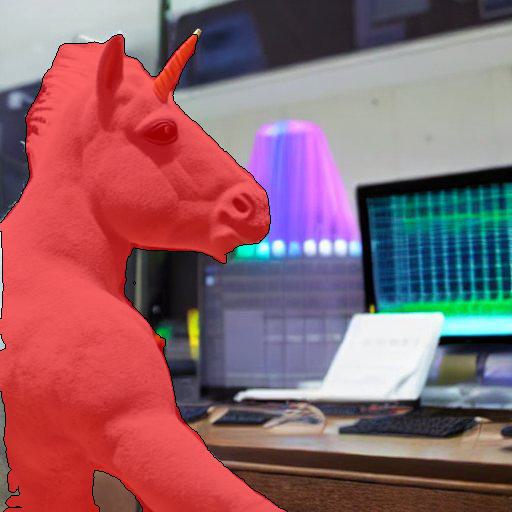}
        \end{subfigure} &
        \begin{subfigure}[b]{0.15\linewidth}
            \includegraphics[width=\textwidth]{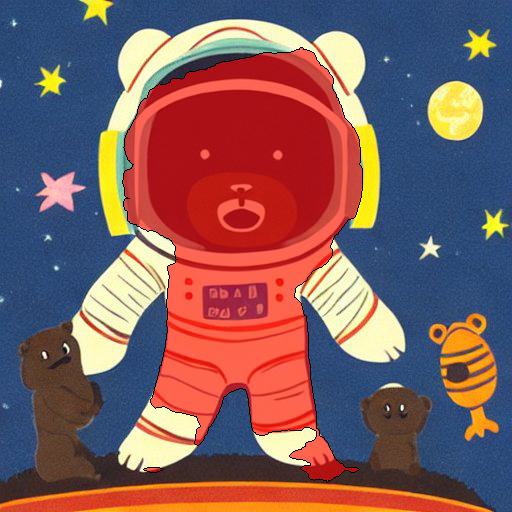}
        \end{subfigure}\\

        \rotatebox{90}{\hskip 3em SeqTR} &
        \begin{subfigure}[b]{0.15\linewidth}
        \includegraphics[width=\textwidth]{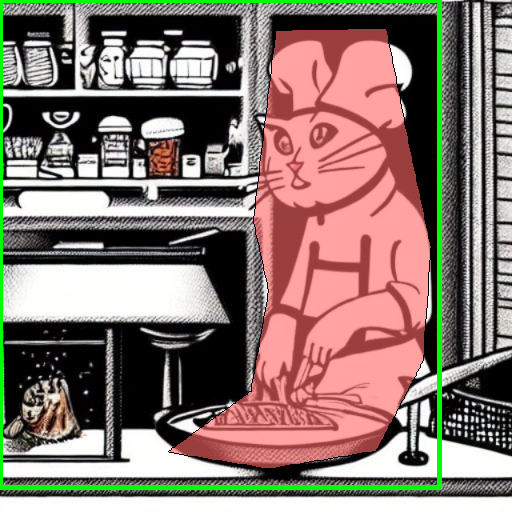}
        \end{subfigure} &
        \begin{subfigure}[b]{0.15\linewidth}
            \includegraphics[width=\textwidth]{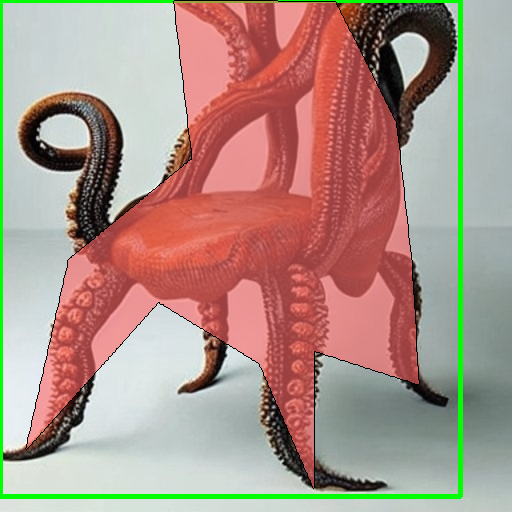}
        \end{subfigure} &  
        \begin{subfigure}[b]{0.15\linewidth}
            \includegraphics[width=\textwidth]{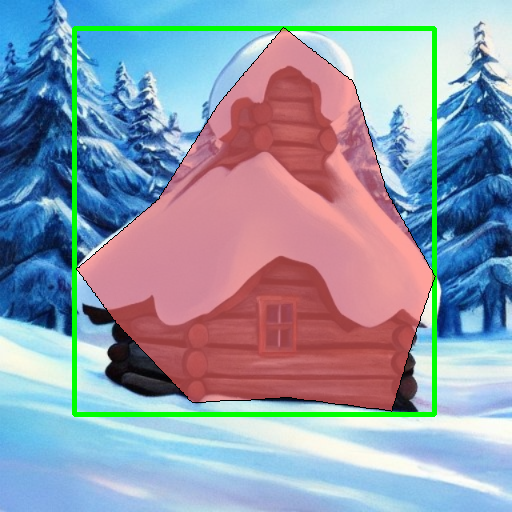}
        \end{subfigure} &
        \begin{subfigure}[b]{0.15\linewidth}
            \includegraphics[width=\textwidth]{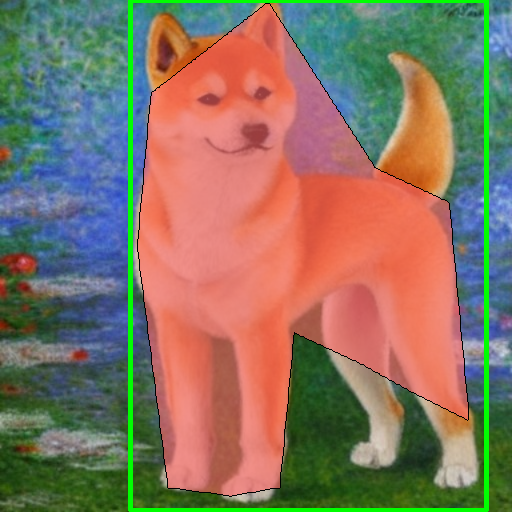}
        \end{subfigure} &
        \begin{subfigure}[b]{0.15\linewidth}
            \includegraphics[width=\textwidth]{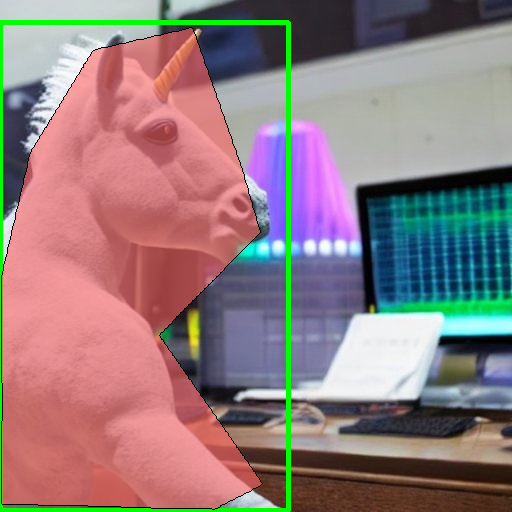}
        \end{subfigure} &
        \begin{subfigure}[b]{0.15\linewidth}
            \includegraphics[width=\textwidth]{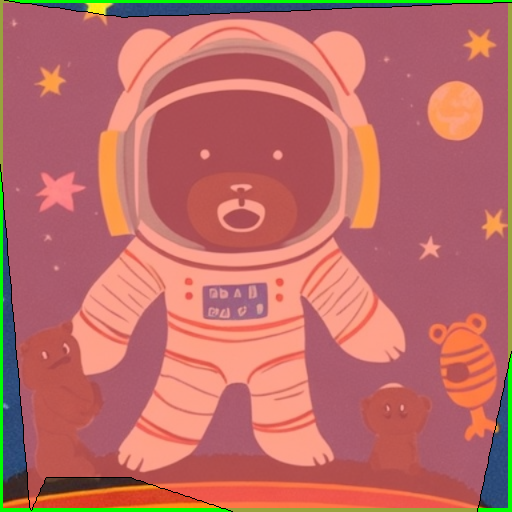}
        \end{subfigure}\\

        \rotatebox{90}{\hskip 1.5em  \textbf{PolyFormer}}&
        \begin{subfigure}[t]{0.15\linewidth}
        \includegraphics[width=\textwidth]{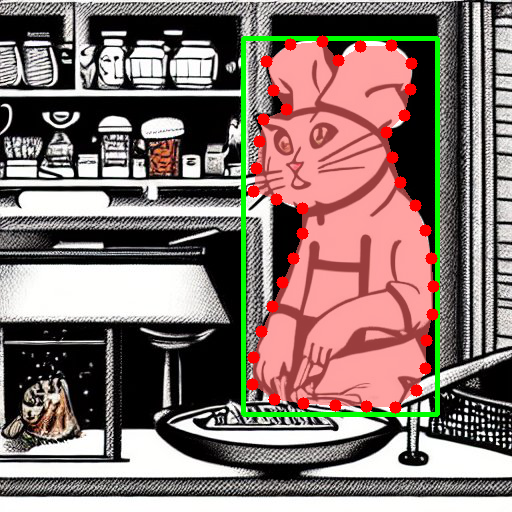}
        \caption{``A cat chef cooking fish in a fancy restaurant"}
        \end{subfigure} &
        \begin{subfigure}[t]{0.15\linewidth}
            \includegraphics[width=\textwidth]{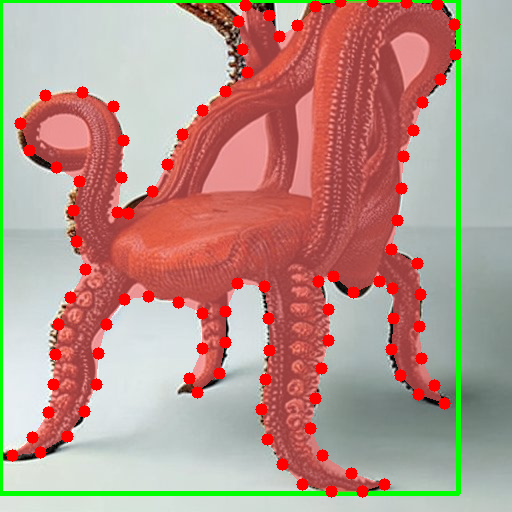}
            \caption{``A chair that looks like octopus"}
        \end{subfigure} &  
        \begin{subfigure}[t]{0.15\linewidth}
            \includegraphics[width=\textwidth]{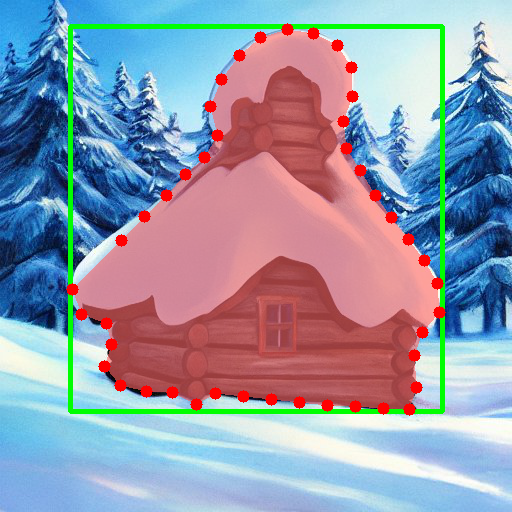}
            \caption{``A small cabin on top of a snowy mountain in the style of Disney artstation"}
        \end{subfigure} &
        \begin{subfigure}[t]{0.15\linewidth}
            \includegraphics[width=\textwidth]{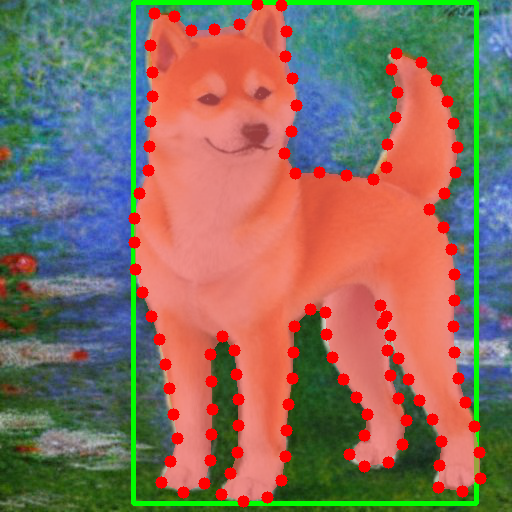}
            \caption{``A shiba inu puppy painted by Monet"}
        \end{subfigure} &
        \begin{subfigure}[t]{0.15\linewidth}
            \includegraphics[width=\textwidth]{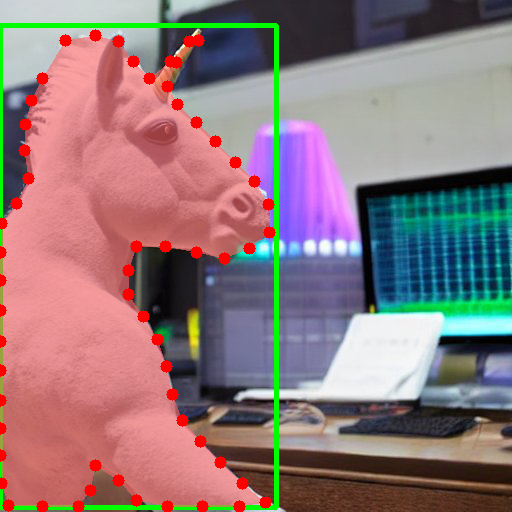}
            \caption{``A unicorn doing computer vision research"}
        \end{subfigure} &
        \begin{subfigure}[t]{0.15\linewidth}
            \includegraphics[width=\textwidth]{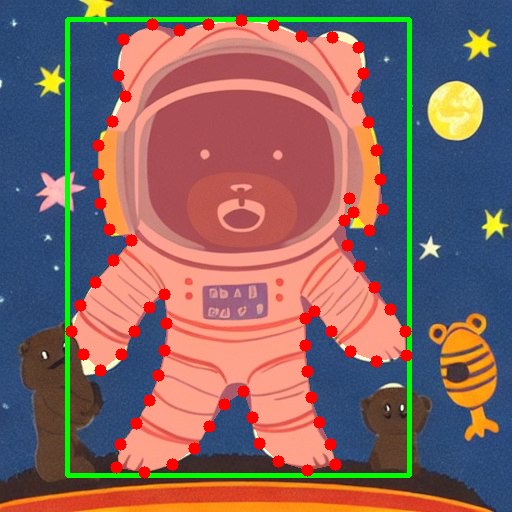}
            \caption{``A bear astronaut in the space"}
        \end{subfigure}\\

        %Input Image & LAVT~\cite{yang2022lavt} & SeqTR~\cite{zhu2022seqtr} & PolyFormer (ours)  &  Ground-truth \\

        % \hline
    \end{tabular}}
    \caption{The result comparison of  LAVT~
    \cite{yang2022lavt}, SeqTR~\cite{zhu2022seqtr} and PolyFormer on synthetic images generated by Stable Diffusion~\cite{rombach2021highresolution}.}
    \label{fig:zero-shot_supp}
    \vspace{10mm}
\end{figure*}

% \begin{tblr}{
%   colspec = {Q[b,4em]|Q[c,m]}
% }
%   LAVT~\cite{yang2022lavt} & \includegraphics[width=0.18\textwidth]{figs/visualization_2/449_1_lavt.png}\\
% \end{tblr}

%% file: figs/visualization_supp.tex
%& $\textrm{XraySyn}_{ref}$ 
\captionsetup[subfigure]{labelformat=empty}
\begin{figure*}[t]
    \setlength{\tabcolsep}{0.5pt}
    \centering
    \small
    \setlength{\tabcolsep}{0.8mm}{\begin{tabular}[b]{lcccccc}
        % \hline
         
        %\multicolumn{5}{l}{Expression: ``a cow that is \textit{\textbf{being fed}} by a bottle"} \\
        \rotatebox{90}{\hskip 3em LAVT} &
        \begin{subfigure}[b]{0.15\linewidth}
        \includegraphics[width=\textwidth]{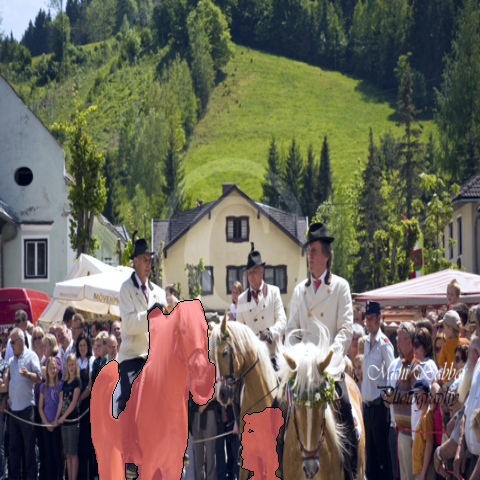}
        \end{subfigure} &
        \begin{subfigure}[b]{0.15\linewidth}
        \includegraphics[width=\textwidth]{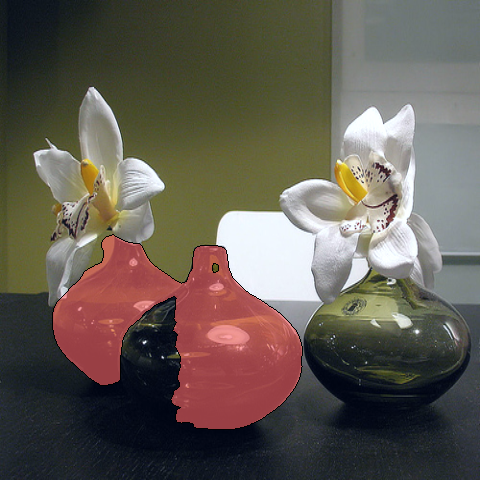}
        \end{subfigure} &
        \begin{subfigure}[b]{0.15\linewidth}
            \includegraphics[width=\textwidth]{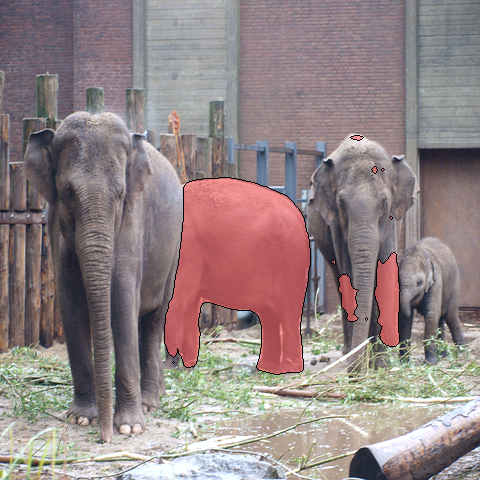}
        \end{subfigure} &  
        \begin{subfigure}[b]{0.15\linewidth}
            \includegraphics[width=\textwidth]{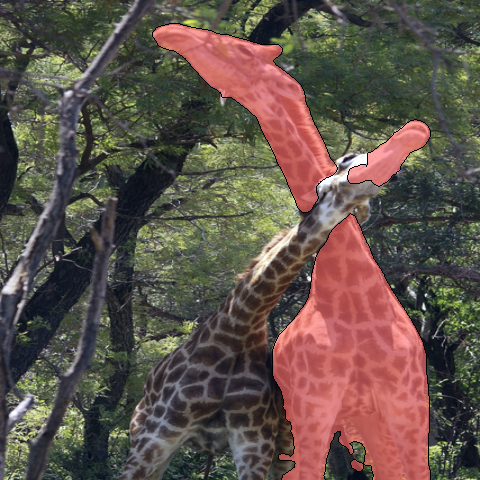}
        \end{subfigure} &
        \begin{subfigure}[b]{0.15\linewidth}
            \includegraphics[width=\textwidth]{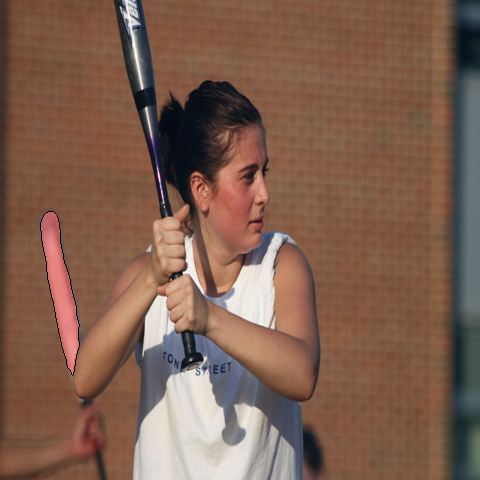}
        \end{subfigure} &
        \begin{subfigure}[b]{0.15\linewidth}
            \includegraphics[width=\textwidth]{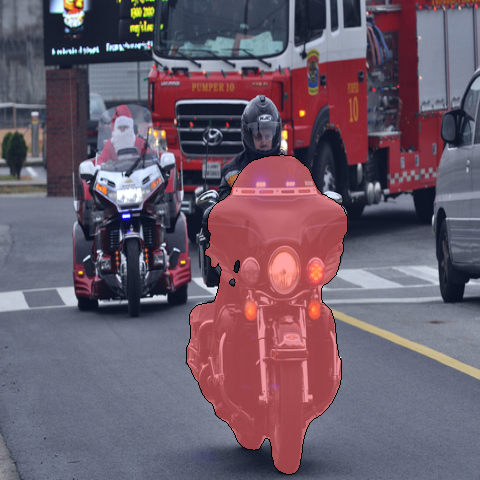}
        \end{subfigure}\\ 
        
        \rotatebox{90}{\hskip 3em SeqTR}&
        \begin{subfigure}[b]{0.15\linewidth}
        \includegraphics[width=\textwidth]{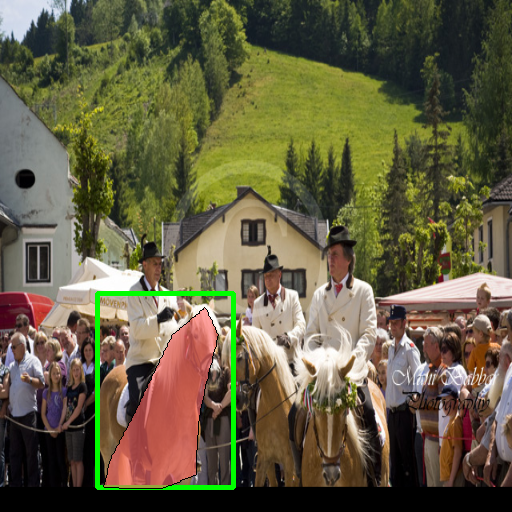}
        \end{subfigure} &
        \begin{subfigure}[b]{0.15\linewidth}
        \includegraphics[width=\textwidth]{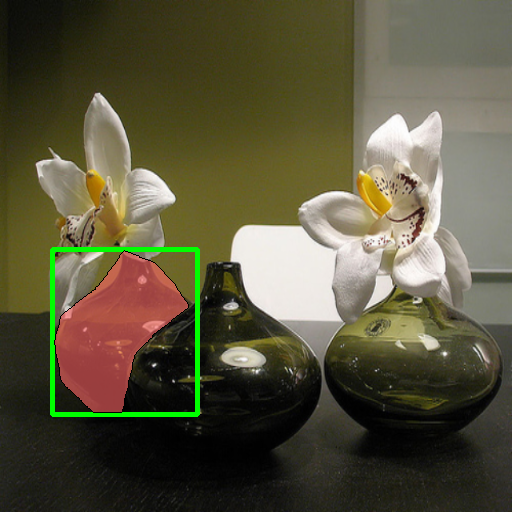}
        \end{subfigure} &
        \begin{subfigure}[b]{0.15\linewidth}
        \includegraphics[width=\textwidth]{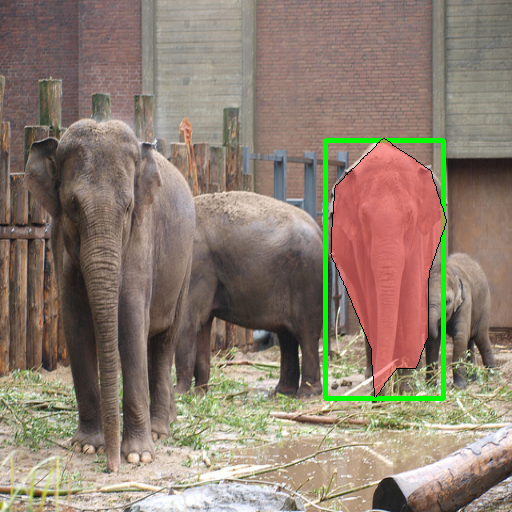}
        \end{subfigure} &  
        \begin{subfigure}[b]{0.15\linewidth}
            \includegraphics[width=\textwidth]{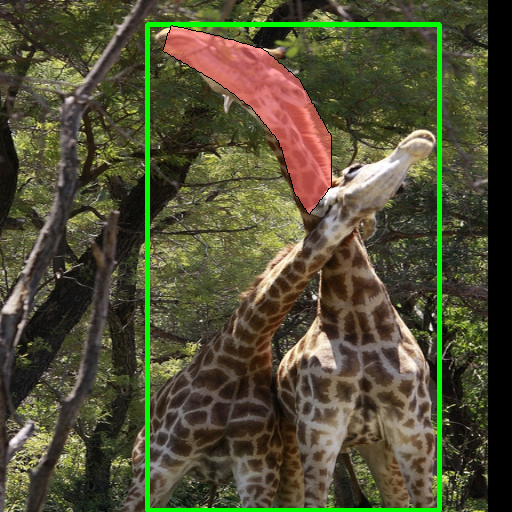}
        \end{subfigure} &
        \begin{subfigure}[b]{0.15\linewidth}
            \includegraphics[width=\textwidth]{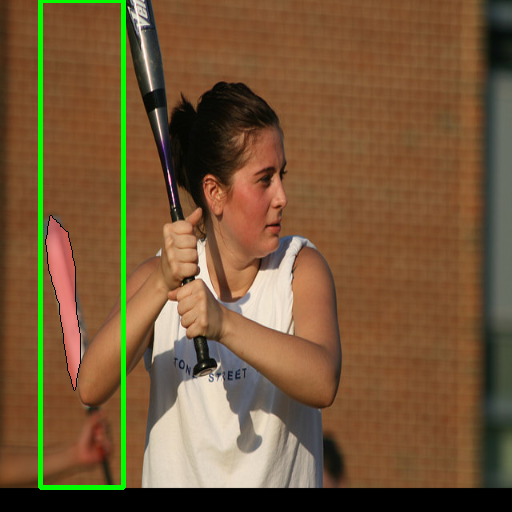}
        \end{subfigure} &
        \begin{subfigure}[b]{0.15\linewidth}
            \includegraphics[width=\textwidth]{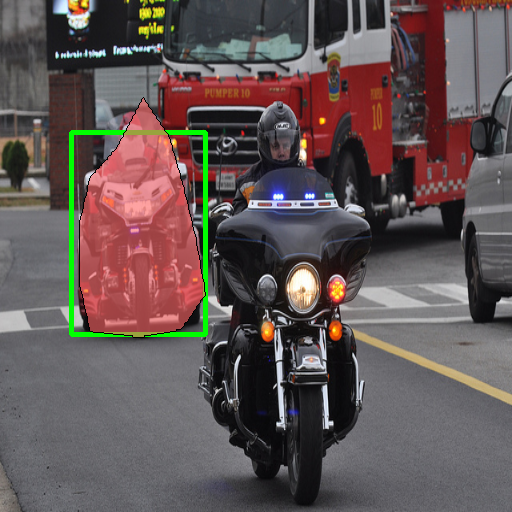}
        \end{subfigure}\\ 
        
        \rotatebox{90}{\hskip 1.5em  \textbf{PolyFormer}}&
        \begin{subfigure}[t]{0.15\linewidth}
        \includegraphics[width=\textwidth]{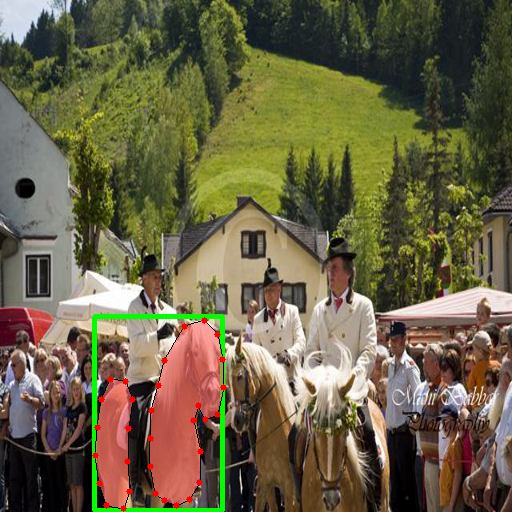}
        \caption{``horse on the left of the group of horses"}
        \end{subfigure} &
        \begin{subfigure}[t]{0.15\linewidth}
        \includegraphics[width=\textwidth]{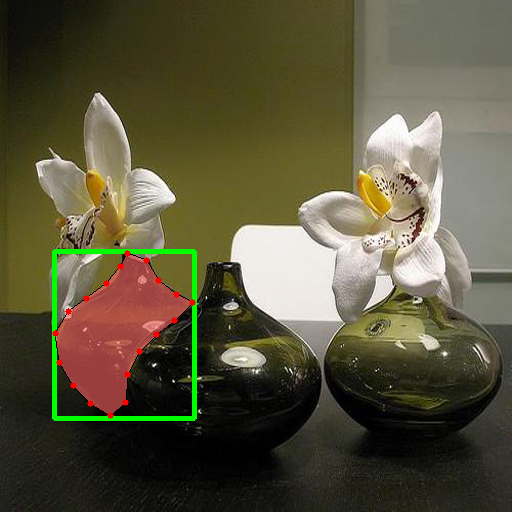}
        \caption{``small green vase on the left with a flower in it"}
        \end{subfigure} & 
        \begin{subfigure}[t]{0.15\linewidth}
            \includegraphics[width=\textwidth]{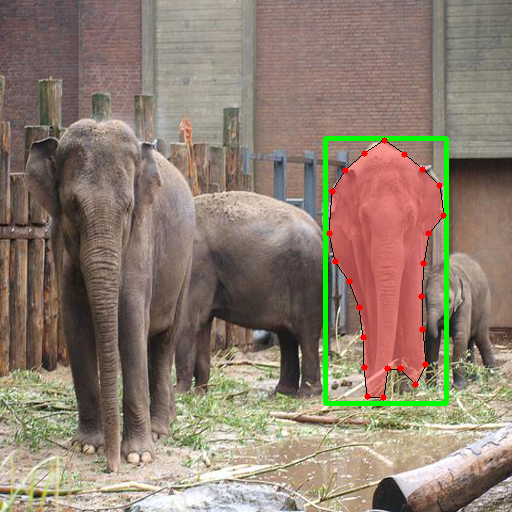}
            \caption{``the elephant with the baby elephant"}
        \end{subfigure} &  
        \begin{subfigure}[t]{0.15\linewidth}
            \includegraphics[width=\textwidth]{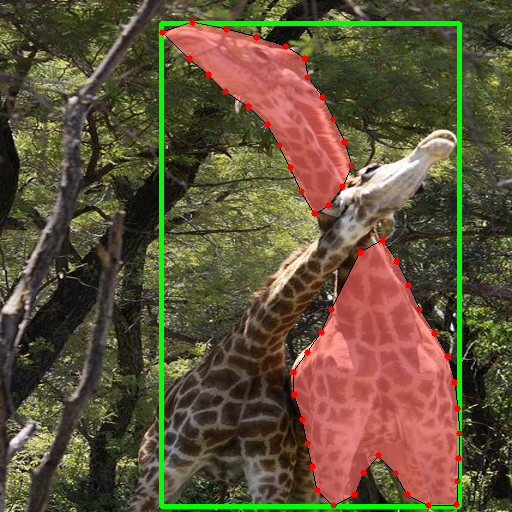}
            \caption{``the taller giraffe"}
        \end{subfigure} &
        \begin{subfigure}[t]{0.15\linewidth}
            \includegraphics[width=\textwidth]{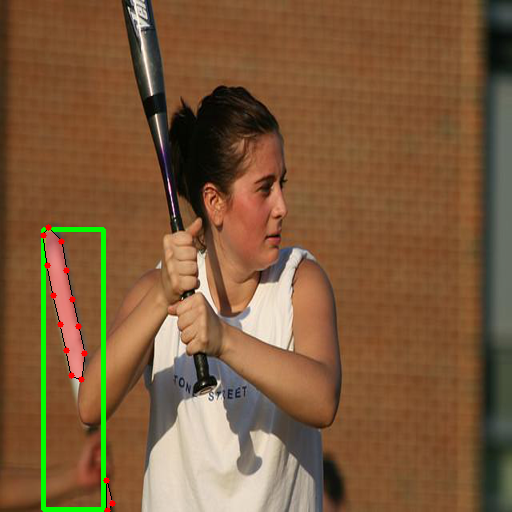}
            \caption{``a white baseball bat, held by a person"}
        \end{subfigure} &
        \begin{subfigure}[t]{0.15\linewidth}
            \includegraphics[width=\textwidth]{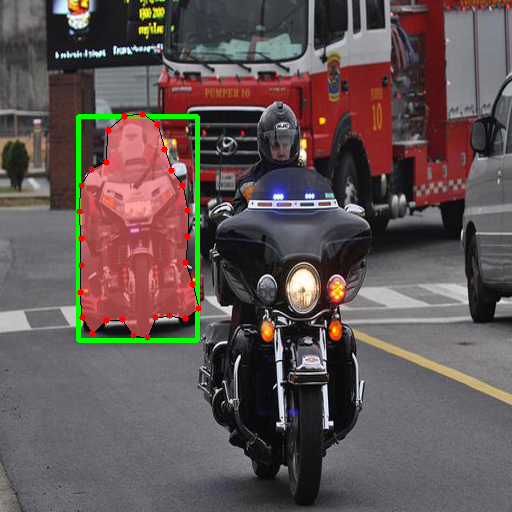}
            \caption{``a red and black motorcycle with a Santa riding it"}
        \end{subfigure} \\ 
        
        \rotatebox{90}{\hskip 3em LAVT} &
        \begin{subfigure}[b]{0.15\linewidth}
        \includegraphics[width=\textwidth]{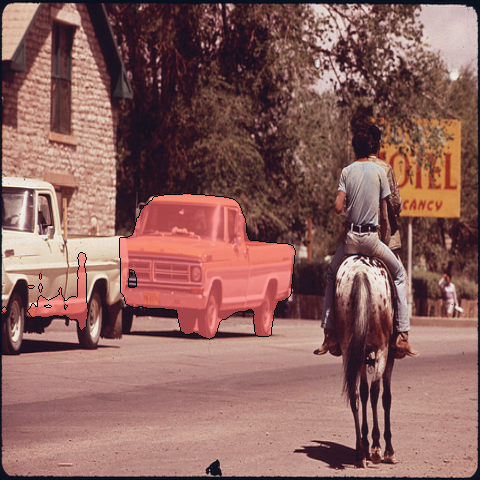}
        \end{subfigure} &
        \begin{subfigure}[b]{0.15\linewidth}
            \includegraphics[width=\textwidth]{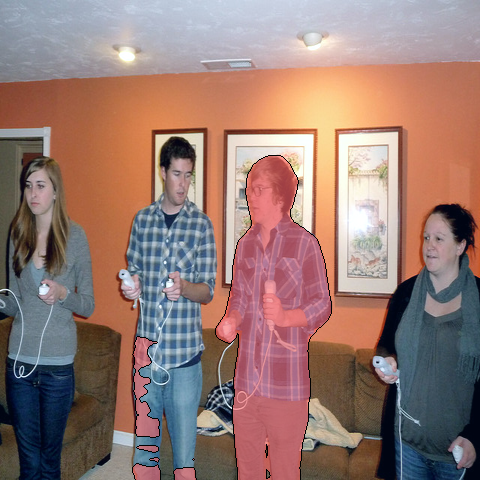}
        \end{subfigure} &  
        \begin{subfigure}[b]{0.15\linewidth}
            \includegraphics[width=\textwidth]{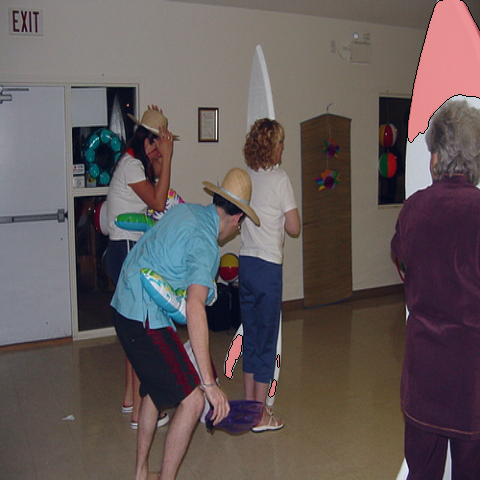}
        \end{subfigure} &
        \begin{subfigure}[b]{0.15\linewidth}
            \includegraphics[width=\textwidth]{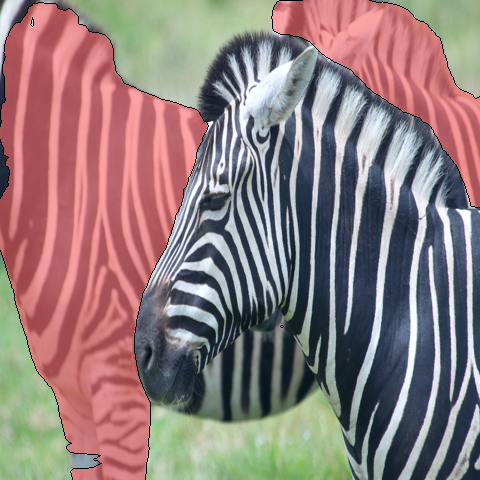}
        \end{subfigure} &
        \begin{subfigure}[b]{0.15\linewidth}
            \includegraphics[width=\textwidth]{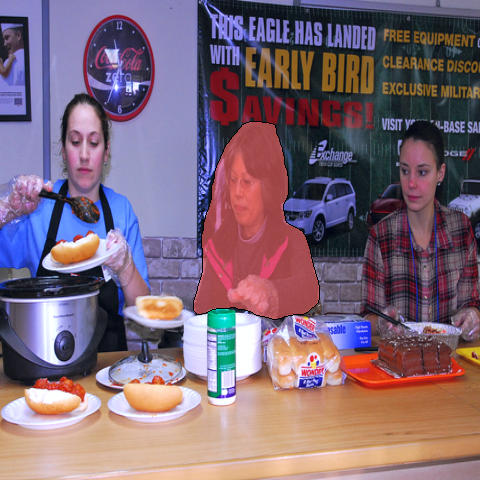}
        \end{subfigure} &
        \begin{subfigure}[b]{0.15\linewidth}
            \includegraphics[width=\textwidth]{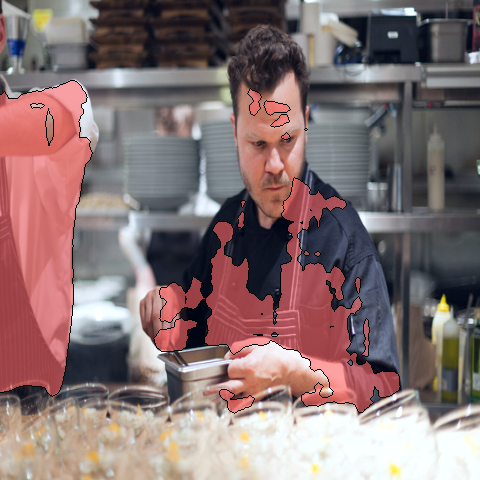}
        \end{subfigure} \\ 
        
        \rotatebox{90}{\hskip 3em SeqTR}&
        \begin{subfigure}[b]{0.15\linewidth}
        \includegraphics[width=\textwidth]{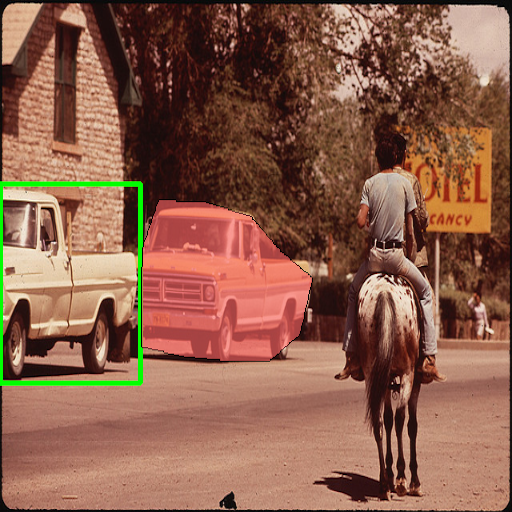}
        \end{subfigure} &
        \begin{subfigure}[b]{0.15\linewidth}
        \includegraphics[width=\textwidth]{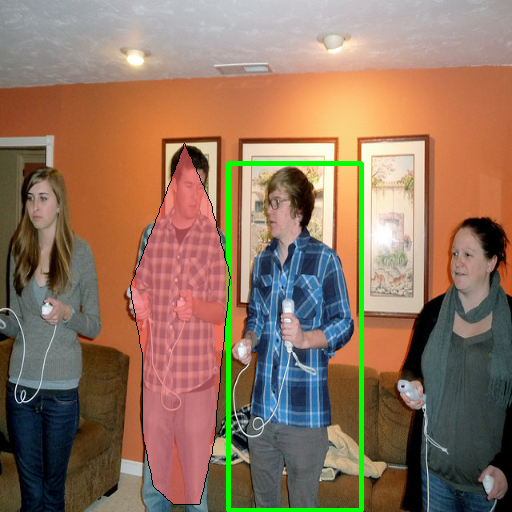}
        \end{subfigure} &  
        \begin{subfigure}[b]{0.15\linewidth}
            \includegraphics[width=\textwidth]{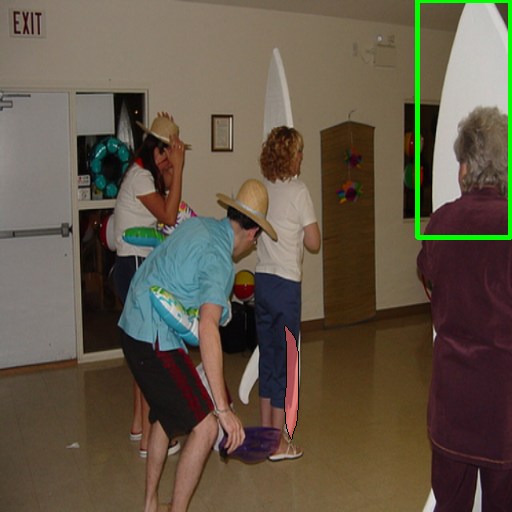}
        \end{subfigure} &
        \begin{subfigure}[b]{0.15\linewidth}
            \includegraphics[width=\textwidth]{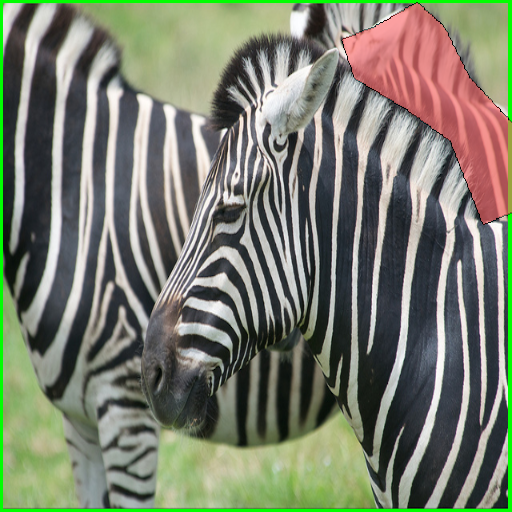}
        \end{subfigure} &
        \begin{subfigure}[b]{0.15\linewidth}
            \includegraphics[width=\textwidth]{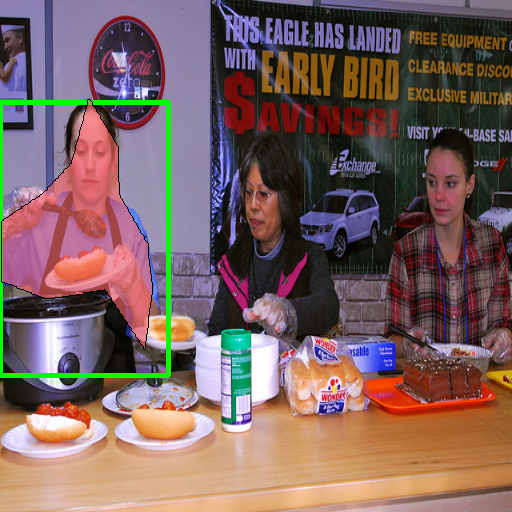}
        \end{subfigure} &
        \begin{subfigure}[b]{0.15\linewidth}
            \includegraphics[width=\textwidth]{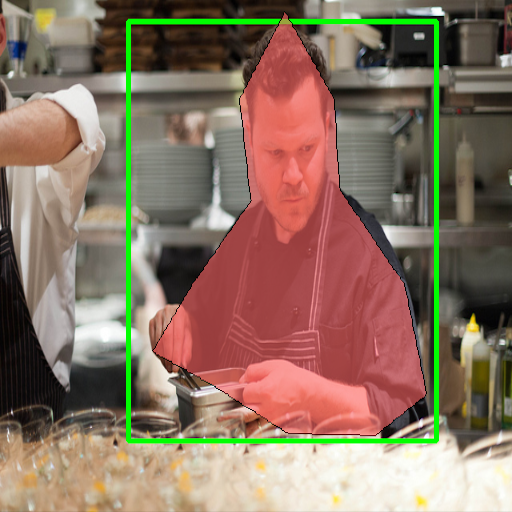}
        \end{subfigure}\\ 
        
        \rotatebox{90}{\hskip 1.5em  \textbf{PolyFormer}}&
        \begin{subfigure}[t]{0.15\linewidth}
        \includegraphics[width=\textwidth]{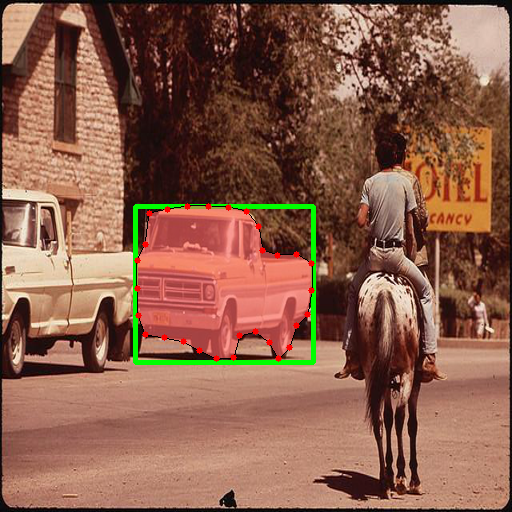}
        \caption{``old yellow and white truck parked behind other truck"}
        \end{subfigure} &
        \begin{subfigure}[t]{0.15\linewidth}
            \includegraphics[width=\textwidth]{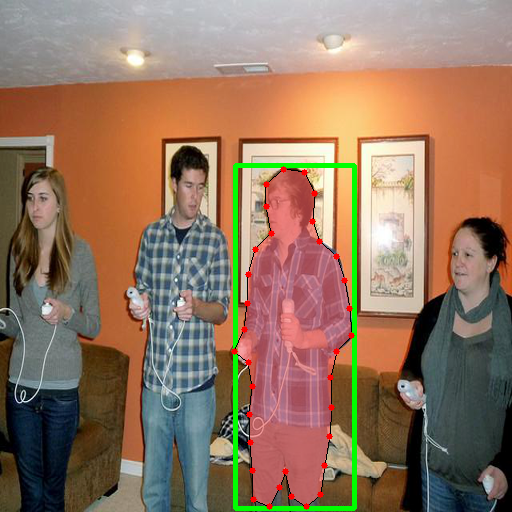}
            \caption{``boy with blue plaid shirt and glasses"}
        \end{subfigure} &  
        \begin{subfigure}[t]{0.15\linewidth}
            \includegraphics[width=\textwidth]{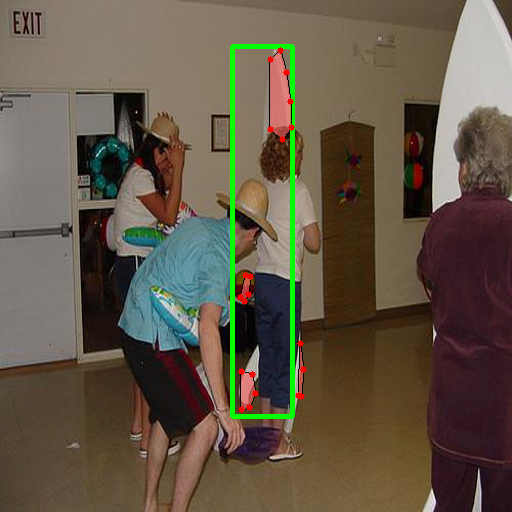}
            \caption{``the surfboard the woman in a white shirt and blue capris is holding"}
        \end{subfigure} &
        \begin{subfigure}[t]{0.15\linewidth}
            \includegraphics[width=\textwidth]{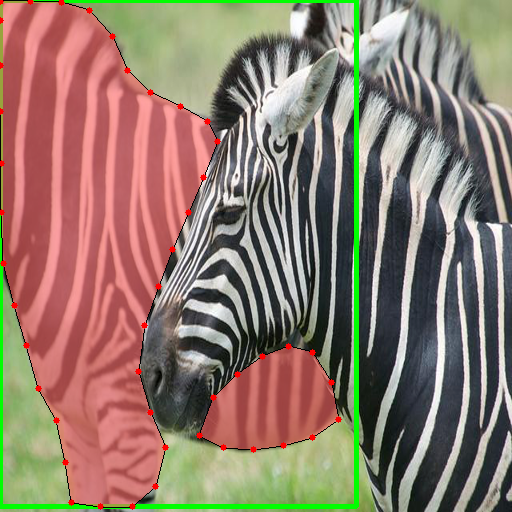}
            \caption{``a zebra with its head not visible but much of its body able to be seen"}
        \end{subfigure} &
        \begin{subfigure}[t]{0.15\linewidth}
            \includegraphics[width=\textwidth]{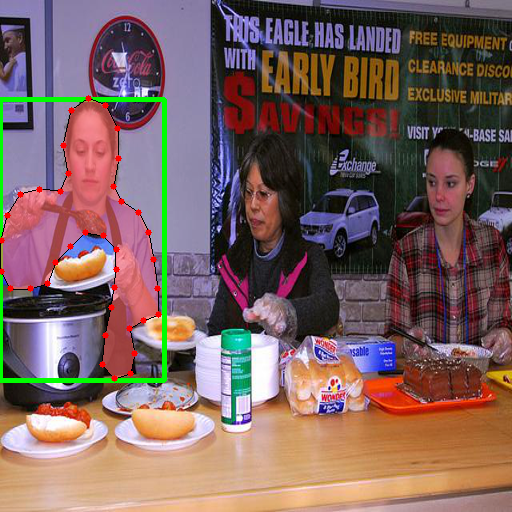}
            \caption{`` a girl was cooking the food and serving"}
        \end{subfigure} & 
        \begin{subfigure}[t]{0.15\linewidth}
            \includegraphics[width=\textwidth]{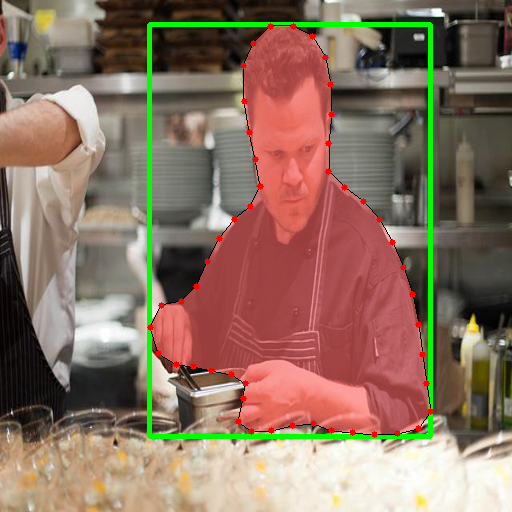}
            \caption{``a man wearing a black shirt and a black and white striped apron stirring something in a metal container"}
        \end{subfigure}\\

        %Input Image & LAVT~\cite{yang2022lavt} & SeqTR~\cite{zhu2022seqtr} & PolyFormer (ours)  &  Ground-truth \\

        % \hline
    \end{tabular}}
    \caption{The result comparison of LAVT~
    \cite{yang2022lavt}, SeqTR~\cite{zhu2022seqtr} and PolyFormer on RefCOCOg test set. PolyFormer simultaneously predicts the bounding box and polygon vertices that forms the segmentation mask. 
    % Blue point indicates the starting point of the vertex sequence. 
    LAVT is for referring image segmentation only. For SeqTR, we generate the bounding boxes and segmentation masks from the task-specific models as they perform better than the multi-task model.}
    % \caption{Visualization results on RefCOCOg. Our unified framework is able to detect and segment the referred object simultaneously. }
    \label{fig:visualization_supp}
    \vspace{10mm}
\end{figure*}